\documentclass[10pt,twocolumn,letterpaper]{article}

\usepackage[pagenumbers]{iccv} 

\usepackage{lineno}
\usepackage{adjustbox}
\usepackage{kotex}
\usepackage{colortbl}
\usepackage{subcaption}
\usepackage{xcolor}
\usepackage{booktabs}
\usepackage{multirow}
\usepackage{pifont}
\newcommand{\cmark}{\ding{51}}
\newcommand{\xmark}{\ding{55}}
\usepackage{amsmath}

\definecolor{iccvblue}{rgb}{0.21,0.49,0.74}
\usepackage[pagebackref,breaklinks,colorlinks,allcolors=iccvblue]{hyperref}

\def\paperID{****} 
\def\confName{ICCV}
\def\confYear{2025}

\title{Superpixel Tokenization for Vision Transformers: \\Preserving Semantic Integrity in Visual Tokens}

\author{
Jaihyun Lew$^{1,*}$
\and
Soohyuk Jang$^{2,*}$
\and
Jaehoon Lee$^{1,*}$
\and
Seungryong Yoo$^2$
\and
Eunji Kim$^2$
\and
Saehyung Lee$^2$
\and
Jisoo Mok$^2$
\and
Siwon Kim$^3$
\and
Sungroh Yoon$^{1,2,4,\dagger}$
\and
$^1$Interdisciplinary Program in AI, Seoul National University \\
$^2$Department of Electrical and Computer Engineering, Seoul National University \\
$^3$Amazon \hspace{15pt}
$^4$AIIS, ASRI, INMC, ISRC, Seoul National University \\
$^*$Equal contribution \hspace{15pt}
$^\dagger$Corresponding Author \\
{\tt\small \{fudojhl, soohyuk.jang, jhcaptain7, sryoon\}@snu.ac.kr} \\
\small \url{https://github.com/jangsoohyuk/SuiT}
}

\newcommand{\ours}{SuiT\xspace}

\begin{document}

\maketitle

\begin{abstract}
Transformers, a groundbreaking architecture proposed for natural language processing (NLP), have also achieved remarkable success in computer vision.
A cornerstone of their success lies in the attention mechanism, which models relationships among tokens.
While the tokenization process in NLP inherently ensures that each token maintains semantic integrity without containing multiple meanings, the grid-based tokenization of Vision Transformer (ViT) relies on uniformly partitioned square image patches, which may result in an arbitrary mixing of visual concepts within a token.
In this work, we propose a novel tokenization pipeline that replaces the grid-based tokenization with superpixels, encouraging each token to capture a distinct visual concept. Unlike square image patches, superpixels are formed in varying shapes, sizes, and locations, making direct substitution challenging. To address this, our pipeline first generates pixel-level embeddings and efficiently aggregates them within superpixel clusters, producing superpixel tokens that seamlessly replace patch tokens in ViT.
Extensive experiments demonstrate that our approach enhances the performance of ViT on various downstream tasks and introduces intriguing properties such as adaptive inference and semantic integrity in tokens.
\end{abstract}

\section{Introduction}

\begin{figure}[!t]
    \centering
    \includegraphics[width=\linewidth]{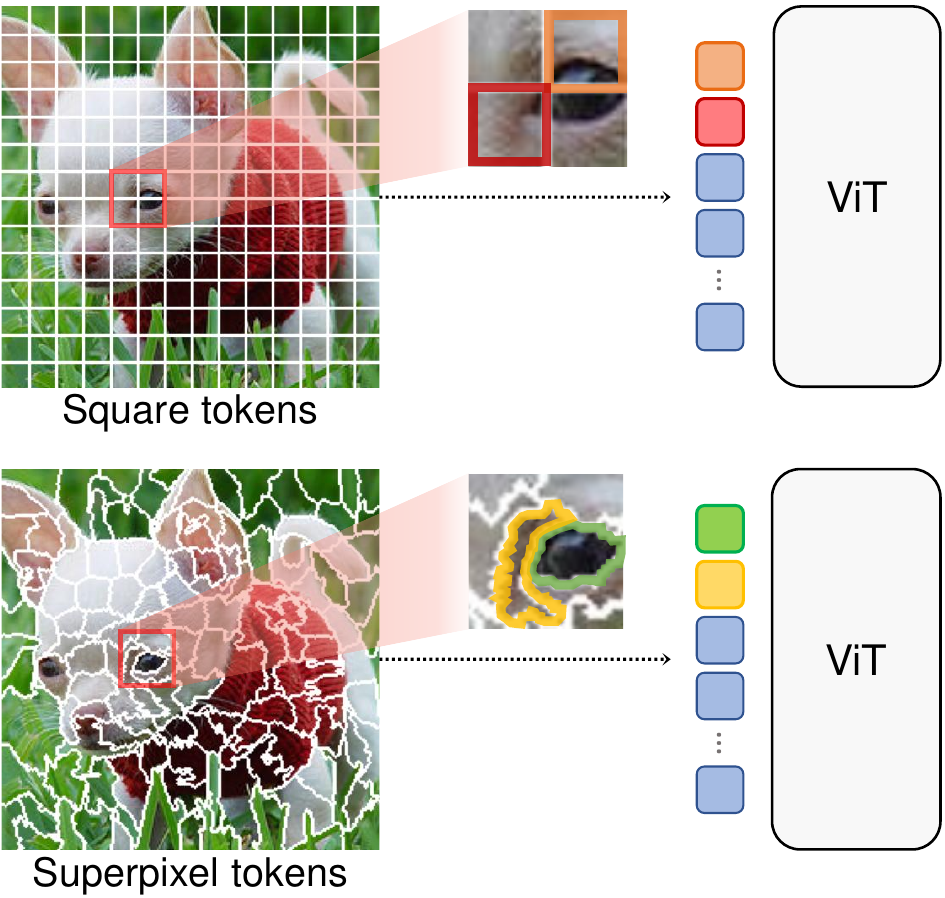}
    \caption{\textbf{A high-level overview of tokenization}. (Top) The conventional grid-like tokenization in ViT and (Bottom) our proposed superpixel tokenization.} 
    \label{fig:thumbnail} 
\end{figure}

Transformers~\cite{transformer}, originally proposed for natural language processing~(NLP), have demonstrated exceptional performance across diverse tasks such as machine translation~\cite{wang2019learning, he2018layer, bapna2018training}, natural language understanding~\cite{kenton2019bert, liu2019roberta}, and text generation~\cite{radfordlanguage, lewis2019bart}.
The transformative power of Transformers was again attested by the release of large language models \cite{gpt3, touvron2023llama}.
Similarly, the Vision Transformer~(ViT)~\cite{vit}, which extended the Transformer architecture to image recognition, changed the landscape of computer vision research.
The subsequent research efforts to broaden the usage of ViTs to various visual tasks, including detection~\cite{shehzadi20232d, detr, deform_detr}, segmentation~\cite{li2024transformer, setr, segmenter}, generation~\cite{vqgan, dit, zhang2022styleswin}, and multi-modal applications~\cite{clip}, have made the ViT the dominant architecture in computer vision.

In NLP, the success of Transformers is commonly attributed to the attention mechanism, which captures contextual dependencies by modeling the relationship among input texts, which are expressed as tokens.
The widely adopted tokenization methods in NLP, such as Byte Pair Encoding~\cite{bytepairencoding}, 
WordPiece~\cite{wordpiece} and SentencePiece~\cite{kudo2018sentencepiece}, first divide each sentence into words
based on whitespaces.
For instance, when tokenizing ``\texttt{John loves Mary},'' each word is separated into: [``\texttt{John}", 
`` ",
``\texttt{loves},"
`` ",
``\texttt{Mary}"].
This initial word-level splitting of sentences prevents multiple words with different meanings from being fused into a single token, preserving semantic integrity in tokens.

In contrast, in computer vision, fixed-size grid patches form the basis of tokenization in ViT and its variants~\cite{vit, deit, swin}.
As visualized in Figure~\ref{fig:thumbnail}, when na\"ively sliced square image patches are used as tokens, a single token can simultaneously contain two or more visual concepts,~\textit{e.g.,} eye and fur.
This is analogous to splitting ``\texttt{John loves Mary}'' into arbitrary fixed-size fragments like [``\texttt{Joh},''  ``\texttt{n l},'' ``\texttt{ove}'', ``\texttt{s M}'', ``\texttt{ary}''], disregarding semantic boundaries.
Consequently, this grid-based tokenization deviates from the semantic integrity inherent in NLP tokenization, leading to tokens with arbitrarily mixed visual concepts.

In this work, we employ the concept of superpixels to design a tokenization process. A superpixel~\cite{slic, seeds, lsc} is a cluster of connected pixels in an image that share similar properties, such as color, texture, or position, and is often used as a unit of semantic representation in various studies~\cite{eac, ace}. Building on this property, superpixels emerge as a viable alternative to square tokens for preserving semantic integrity.
Figure~\ref{fig:thumbnail} illustrates how leveraging superpixels alters tokenization of ViT.
Grouping pixels into superpixels divides an image into cohesive regions that capture similar semantics, thereby minimizing the mixture of unrelated semantic information in each token.

Despite the potential advantages of superpixels in tokenization, the different shapes and sizes of each superpixel present significant challenges in utilizing them as a direct replacement for visual tokens~\cite{ssn, supfcn, proxy}. In ViTs, tokens are generated from fixed-size square patches at predefined locations through simple flattening and linear projection operations.
Because superpixels consist of varying numbers of pixels, \textit{de facto} flattening and linear projection operations in ViT are incompatible with superpixels.
Moreover, the location of each superpixel changes dynamically depending on the image.
The variability of superpixel locations makes it infeasible to directly apply the positional embeddings of ViT, which are designed for fixed locations. 
The inadequacy of original ViT operations necessitates novel techniques to address irregular geometry and composition to reap the inherent benefits of superpixels.

We thus propose a novel tokenization pipeline that overcomes the aforementioned problems and enables the effective incorporation of superpixels into ViT tokenization.
The proposed pipeline consists of two major technical components:
\textbf{pixel-level embedding} and \textbf{superpixel-aware aggregation}.
The first stage generates pixel-level embeddings consisting of local and positional features. Second, these pixel-level embeddings are then aggregated according to superpixel clusters. The aggregation process employs pooling operations to remove the irregularity and variability of superpixels.
Collectively, this two-stage process allows the proposed tokenization approach to leverage the benefits of superpixels while preserving important details of an image.

We demonstrate the superiority of our \textbf{Su}perpixel-Tokenized V\textbf{i}sion \textbf{T}ransformer (\ours) compared to other baselines. \ours outperforms in a variety of tasks including ImageNet-1K classification \cite{imagenet}, transfer learning \cite{transferlearning}, and zero-shot segmentation \cite{tokencut}. We further analyze intriguing properties that emerge using our tokenization pipeline, such as adaptive inference and semantic integrity.

Our contribution can be summarized as follows:
\label{sec:intro}

\begin{itemize}[leftmargin=*, nolistsep]
    \item{We propose an effective superpixel-based tokenization pipeline that overcomes the challenges associated with the use of superpixels.}
    \item{We empirically show that our tokenization pipeline induces notable properties including adaptive inference and semantic integrity in tokens.}
    \item{We demonstrate the effectiveness of our approach through experiments on image classification, transfer learning, and zero-shot segmentation.}
\end{itemize}
\section{Related Work}

\paragraph{Superpixel-based Vision Models}
Superpixels, which are perceptually meaningful clusters of pixels, have been a foundational concept in computer vision~\cite{slic,seeds,lsc}. 
Previous studies have attempted to use superpixels in several tasks such as segmentation~\cite{kwak2017weakly, spformer} and object detection~\cite{he2015supercnn}.

A recent work, STViT~\cite{supertoken}, applies superpixel-like clustering named supertoken attention in the Transformer backbone. However, unlike superpixels which are formed at the pixel level, clusters in STViT are formed at the square token level, resulting in coarser granularity.
CoC~\cite{coc} borrows the concept of superpixels within neural networks, but does not actually exploit superpixels in the network.
Similarly, SPFormer~\cite{spformer} utilizes superpixel-based representations but requires specialized attention modules making it difficult to integrate with existing ViT architectures in a plug-and-play manner.

These works all modify the transformer backbone to incorporate the concept of superpixels but do not explore their use in tokenization. In contrast, our work introduces a simple yet effective superpixel tokenization method that can be seamlessly integrated into the Transformer architecture without altering the backbone for a range of general-purpose tasks. 

\paragraph{Tokenization Methods}
While most ViT variants focus on improving backbone architectures and attention mechanisms using na\"ive square image patches, there have been several attempts in exploration of advanced tokenization strategies tailored for vision tasks. Quadformer~\cite{quadformer} and MSViT~\cite{msvit} both propose adaptive tokenization strategies that dynamically adjust token resolutions based on the image content. 

The studies SPiT~\cite{spit} and sViT~\cite{svit} are the most similar to our work as they focus solely on the tokenization module while keeping the backbone intact and share similar research motivations. SPiT~\cite{spit} uses superpixel tokenization but focuses on analyzing the tokenization from the perspective of explainable AI but do not demonstrate favorable performance compared to original ViT. 

sViT~\cite{svit} shares a similar motivation in preserving semantic integrity during tokenization, but it still divides tokens based on a na\"ive bounding box, failing to create tokens with semantic integrity. Additionally, resizing these tokens into square shapes compromised the visual structure of objects within the tokens. These shortcomings in design leads to limited performance in various tasks.

As described in Section~\ref{sec:intro}, we propose a tokenization that encourages semantic integrity in tokens through superpixels. We investigate the performance of our method across various tasks and explore its intriguing properties.

\begin{figure*}[!ht]
    \centering
    \includegraphics[width=\linewidth]{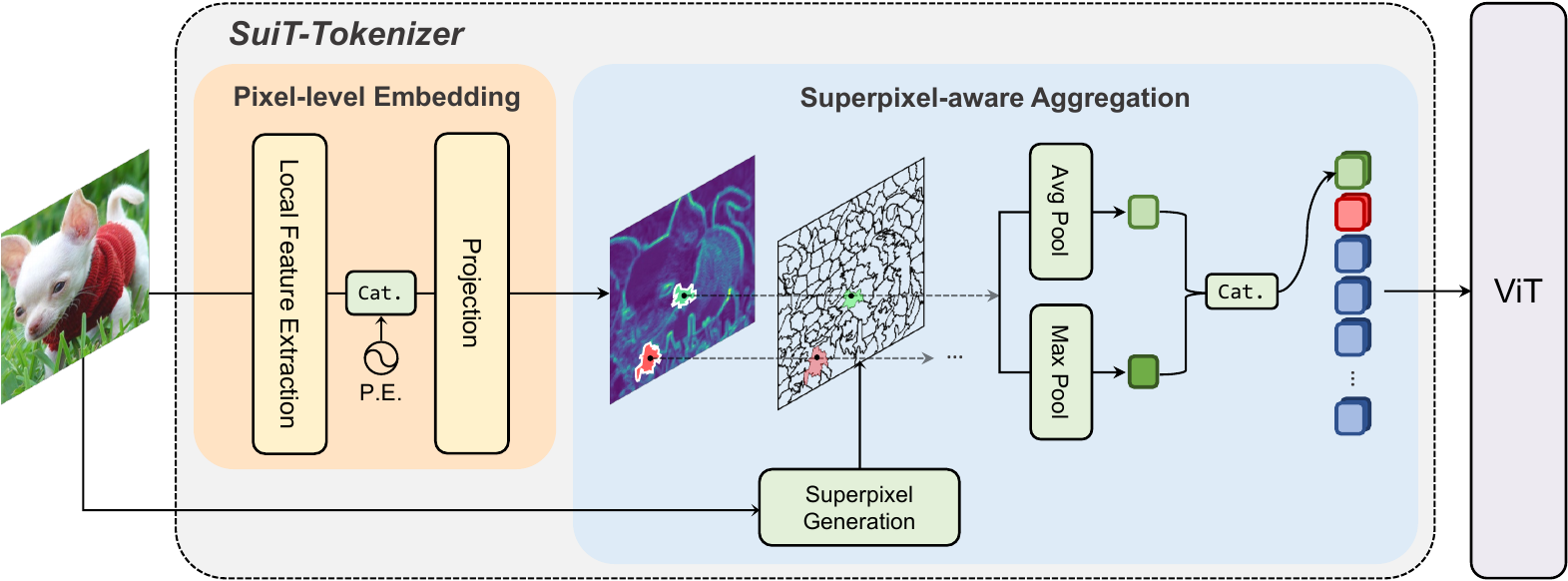} 
    \caption{\textbf{Overview of our superpixel tokenization pipeline.} Local features are extracted and combined with positional encodings, followed by superpixel-aware aggregation using average and max pooling to produce superpixel tokens, which are fed into Vision Transformer.} 
    \label{fig:main_figure} 
\end{figure*}

\section{Preliminaries}
\label{sec:preliminaries}
In this section, we introduce the key concepts and notations that are integral to our work, which aims to construct ViT tokens that preserve semantic integrity using superpixels.
In Section~\ref{sec:superpixel}, we describe the mechanism of superpixel algorithms, and in Section~\ref{sec:token_in_vit}, we go over the traditional ViT tokenization.
Subsequently, we discuss the research challenges in adopting superpixels as tokens in ViTs.

\subsection{Superpixel Algorithms}
\label{sec:superpixel}
Superpixels are clusters of image pixels grouped together based on color, position, or other similarity measures.~\cite{slic, seeds, lsc}
They provide a compact and meaningful representation of the image by grouping pixels into regions that align with its underlying structure.

Given an input image $\mathbf{x} \in \mathbb{R}^{H \times W \times C}$, a superpixel algorithm $\mathcal{S}$ produces a superpixel index map $\mathcal{I} \in \mathbb{R}^{H \times W \times 1}$:
\begin{equation}
    \mathcal{I} = \mathcal{S}(\mathbf{x}, K, m),
\end{equation}
where $K$ is the number of superpixels, and $m$ is a parameter that controls the trade-off between color and positional distance, \textit{i.e.}, compactness in SLIC~\cite{slic}.

The superpixel index of at coordinate $(h,w)$ can be defined as 
$\mathcal{I}^{(h,w)}$, where  $0< h \leq  H,  0 < w \leq W$. 
A $k$-th superpixel $\mathcal{C}_k$ can then be defined as:
\begin{equation}
    \mathcal{C}_k = \{\textbf{x}^{(h,w)} \mid \mathcal{I}^{(h,w)}=k, 0 < k \leq K\},
\end{equation}
where $\textbf{x}^{(h,w)}$ is the pixel at $(h,w)$. 
Each superpixel $\mathcal{C}_k$ consists of varying numbers of pixels with irregular spatial distributions.

\subsection{Tokenization in Vision Transformers}
\label{sec:token_in_vit}
In grid-based tokenization of conventional ViT and its variants~\cite{deit, swin, clip}, the input image $\mathbf{x}$ is divided into \(N = \frac{H \times W}{P^2}\) non-overlapping patches of size \(P \times P\), each represented as \(\mathbf{p}^{(i)} \in \mathbb{R}^{P \times P \times C}\), where $\mathit{i}\in\{1, ..., N\}$ corresponds to the index of each patch. Each patch \(\mathbf{p}^{(i)}\) is flattened into a vector $\mathbf{p}_{flat}^{(i)} \in \mathbb{R}^{P^2C}$ and linearly projected using an embedding matrix \(\mathbf{E} \in \mathbb{R}^{P^2C \times D}\), where $D$ is the size of hidden dimension.
In addition to this, positional embedding $\mathbf{E}^{(i)}_{pos} \in \mathbb{R}^D$ of its location is injected to form the input token \( \mathbf{z}^{(i)} = \mathbf{p}_{flat}^{(i)} \mathbf{E} +\mathbf{E}^{(i)}_{pos} \in \mathbb{R}^D \).
The set of tokens $\mathbf{Z} = \{\mathbf{z}^{(i)} \mid i \in \{1,2, ..., N\} \}$ which originates from patches as above, is given to the ViT models along with a learnable classification token $\mathbf{z}_{\text{cls}}$ for classification.

\paragraph{Challenges}
The tokenization process in ViT requires square-shaped and fixed-sized patches, acquired at uniform locations.
These constraints on image patches make a na\"ive application of superpixels to ViT tokenization challenging.
First, the straightforward flatten-and-project operation is incompatible with superpixels that come in varying sizes and shapes~\cite{ssn, supfcn, proxy}.
Flattening of superpixels with diverse shapes and varying numbers of pixels yields vectors of varying lengths, which hinder a simple linear projection.
Second, unlike grid patches with fixed positions, superpixels are located at different positions depending on the images.
Consequently, the standard positional embedding, which relies on fixed positions, is inappropriate for superpixels.

\section{Method}
\label{sec:method}
To address the issues in Section~\ref{sec:preliminaries}, we propose a superpixel tokenization pipeline that embeds each superpixel into a single token.
The overview of superpixel tokenization is illustrated in Figure~\ref{fig:main_figure}.
Our pipeline consists of two components: \textbf{Pixel-level Embedding}~(Section~\ref{subsec:feat_ext}) and \textbf{Superpixel-aware Aggregation}~(Section~\ref{subsec:feat_agg}).
Our aggregation is based on pooling operations to address the varying sizes and shapes of superpixels.
However, na\"ive application of such aggregation on the RGB space may lead to critical loss of information. (Table~\ref{tab:ablation_feats})
To sidestep this potential risk, we use a pixel-level embedding layer to enrich pixel-level information before superpixel-aware aggregation.
These two processes of the proposed tokenization method together exploit the benefits of superpixel-based tokenization while conserving critical image information.

\subsection{Pixel-level Embedding}
\label{subsec:feat_ext}

\paragraph{Local Features}
We start by extracting local features from the input image using a simple convolutional block.
We adopt the design of the initial convolutional block in ResNet~\cite{resnet},
which consists of a $7 \times 7$ convolutional layer, a Batch Normalization layer~\cite{batchnorm}, and a ReLU activation function~\cite{relu}.
We make a slight modification by replacing ReLU with GELU~\cite{gelu}, to align with recent advances in the field~\cite{vit, swin}.
Given an input image $\mathbf{x}$ and the convolutional block, we extract local features $\mathbf{F}_{local}\in\mathbb{R}^{H \times W \times d_{local}}$.

\paragraph{Positional Features}
Positional features play a crucial role in Transformers, as they are unaware of the order of input tokens.
While the original Transformer proposed to use sinusoidal positional encodings (PE) with pre-defined frequencies~\cite{transformer}, ViTs opted to use learnable positional embeddings for each pre-defined locations~\cite{vit, deit}.

Superpixels have complex positional information of high granularity. Using the learnable positional embeddings along every pixels as in ViT is inadequate, as it would require excessive number of parameters, \textit{i.e.,} $H \times W$ learnable embeddings each of $D$-dimensions. 
To keep it parameter-efficient, we adopt the sinusoidal positional encoding with learnable frequencies, which is known to capture high frequency information efficiently~\cite{fourierfeatures}.
To our knowledge, this positional encoding scheme has not been applied to the Transformer literature before, and is a peculiar choice to adapt to our design of using superpixels of high-granularity.

Consequently, we acquire the value of $q$-th dimension in $d_{pos}$-dimensional positional feature $\mathbf{F}_{pos} \in \mathbb{R}^{H\times W \times d_{pos}}$ at each spatial coordinate $(h,w)$ as below:
\begin{equation}
    \begin{split}
    &\mathbf{F}^{(h,w)}_{pos}[2q] = \sin(f_y[q] \cdot h + f_x[q] \cdot w), \\
    &\mathbf{F}^{(h,w)}_{pos}[2q+1] = \cos(f_y[q] \cdot h + f_x[q] \cdot w),
    \end{split}
\end{equation}
where $f_x[q]$ and $f_y[q]$ each denote the frequency of horizontal and vertical axis in the $q$-th dimension.

\paragraph{Projection}
Given the local features $\mathbf{F}_{local}$ and the positional features $\mathbf{F}_{pos}$,
we integrate the two to obtain our pixel-level embeddings.
The two feature maps $\mathbf{F}_{local}$ and $\mathbf{F}_{pos}$ are concatenated and passed through a linear projection layer, represented as a weight matrix $\mathbf{W}_{P}$:
\begin{equation}
    \mathbf{F} = 
    \langle\mathbf{F}_{local}, \mathbf{F}_{pos}\rangle\mathbf{W}_{P}
\end{equation}
where $\langle\cdot\rangle$ denotes concatenation in the channel dimension and $\mathbf{W}_P \in \mathbb{R}^{(d_{local}+d_{pos}) \times \frac{D}{2}}$ is the projection matrix. The resulting pixel-level embeddings $\mathbf{F}\in \mathbb{R}^{H\times W \times \frac{D}{2}}$ is used in the subsequent aggregation stage.

\subsection{Superpixel-aware Aggregation} \label{subsec:feat_agg}
To form an embedding of each superpixel, we aggregate the pixel-level embeddings within each superpixel.
The superpixel index map $\mathcal{I}$, acquired by the superpixel algorithm, serves as a guideline in determining which embeddings correspond to which superpixels.
Based on $\mathcal{I}$, we aggregate $D/2$-dimensional pixel-level embeddings within each superpixel via average and max pooling, which effectively handles embeddings of varying numbers.
This results in two embedding vectors per superpixel: average embeddings $\mathbf{z}_{\text{avg}}$ and max embeddings $\mathbf{z}_{\text{max}}$, each of $D / 2$-dimensions: 
\begin{equation}
\begin{split}
    \mathbf{z}^{(k)}_{\text{avg}} &= \frac{1}{|\mathcal{C}_k|} \sum_{\textbf{x}^{(h,w)} \in \mathcal{C}_k} \mathbf{F}^{(h,w)}, \\
    \mathbf{z}^{(k)}_{\text{max}} &= \max_{\textbf{x}^{(h,w)} \in \mathcal{C}_k} \mathbf{F}^{(h,w)},
\end{split}
\end{equation}
where $|\mathcal{C}_k|$ and $\mathbf{F}^{(h,w)}$ denote the number of pixels within superpixel $\mathcal{C}_k$ and the pixel-level embeddings at coordinate $(h, w)$, respectively.

Average pooling and max pooling extract complementary embeddings from each superpixel. (\ref{tab:ablation_agg}) Average pooling captures the overall characteristics common in the superpixel, while max pooling identifies the most salient features that may be obscured by averaging alone.
We concatenate $D/2$-dimensional vectors from each pooling method along the channel dimension, yielding a $D$-dimensional token $\mathbf{z}^{(k)}$:
\begin{equation}
    \mathbf{z}^{(k)} = \langle \mathbf{z}^{(k)}_{\text{avg}}, \mathbf{z}^{(k)}_{\text{max}}\rangle.
\end{equation}
This approach effectively combines both general and prominent information, creating a more comprehensive representation of each superpixel.
\section{Experiments}

\begin{figure*}[t!]
  \centering
  \small
  \begin{minipage}[t]{0.6\textwidth}
    \vspace{5pt} 
    \centering
    \setlength{\tabcolsep}{2.5pt}
    \begin{tabular}{c@{\hspace{10pt}}l c c c c c}
    \toprule
         Size & Model & Weight-Init. & \# Params. & GMACs &  Top-1 (\%) \\
    \midrule\midrule
        \multirow{6}{*}{Tiny}   & DeiT~\cite{deit} & \multirow{3}{*}{Random} & 5.7M & 1.26  & 72.2\\
                                & SPiT~\cite{spit} &  & - & -  & - \\
                                &\ours &  & 5.5M & 1.44  & \textbf{75.7} \\
        \cmidrule{2-6}
                                & MSViT~\cite{msvit} & \multirow{3}{*}{IN-21K} & - & 0.98 & 75.5 \\
                                & Quadformer$^\dagger$~\cite{quadformer} &  & - & - & - \\
                                & \ours$^\dagger$ &   & 5.5M & 1.44 & \textbf{76.1} \\
                                
    \midrule
    
        \multirow{6}{*}{Small} & DeiT~\cite{deit} & \multirow{3}{*}{Random} & 22.1M & 4.61 & 79.8\\
                                & SPiT~\cite{spit} &  & - & -  & 75.0 \\
                                & \ours &  & 21.7M & 5.20  & \textbf{80.9} \\
        \cmidrule{2-6}
                                & MSViT~\cite{msvit} & \multirow{3}{*}{IN-21K} & - & 4.43 & 82.0 \\
                                & Quadformer$^\dagger$~\cite{quadformer} &  & - & 4.71 & 80.8 \\
                                & \ours$^\dagger$ &  & 21.7M & 5.20 & \textbf{82.6} \\
                                
    \midrule
    
        \multirow{6}{*}{Base} & DeiT~\cite{deit} & \multirow{3}{*}{Random} & 86.8M & 17.6 & 81.8 \\
                                & SPiT~\cite{spit} &  & - & - & 80.4 \\
                                & \ours &  & 86.0M & 19.7  & \textbf{82.3} \\
        \cmidrule{2-6}
                                & MSViT~\cite{msvit} & \multirow{3}{*}{IN-21K} & - & - & - \\
                                & Quadformer$^\dagger$~\cite{quadformer} &   & - & 17.7 & 84.4 \\
                                & \ours$^\dagger$ &   & 86.0M & 19.7 & \textbf{84.5} \\
     \bottomrule
    \end{tabular}
    \captionof{table}{\textbf{Comparison of classification performance on ImageNet-1K~\cite{imagenet}.}
    Following prior studies, we condcuted experiments on two different settings for fair comparisons.
    In the ``Weight-Init." column, ``Random" refers to from-scratch training on ImageNet-1K~\cite{imagenet}, while ``IN-21K" 
    refers to training from ViT weights pre-trained on ImageNet-21K~\cite{imagenet_21k}. $^\dagger$ denotes that the method was initialized with IN-21K pre-trained + IN-1K fine-tuned ViT weights. Cells marked `-' means that the value is not available.
    }
    \label{tab:main_table}
    \end{minipage}
  \hfill
  \begin{minipage}[t]{0.38\textwidth}
    \vspace{0pt} 
    \centering
    \includegraphics[width=\linewidth]{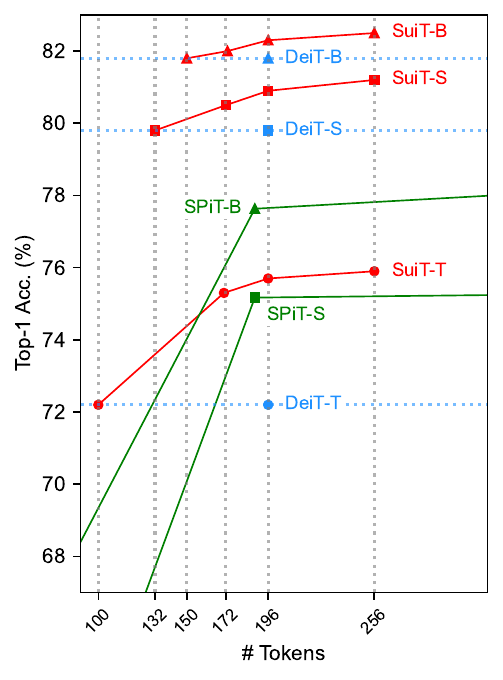}
    \captionof{figure}{\textbf{Adaptive inference.} Under the same image resolution (=224), DeiT~\cite{deit} is forced to use a fixed \# tokens (=196), whereas both \ours and SPiT~\cite{spit} can adaptively adjust the number of tokens during inference. Notably, across all token counts and model scales, \ours consistently achieves the best performance compared to baselines under the same computational cost.}
\label{fig:adaptive_inference}
  \end{minipage}
  \vspace{-1em}
\end{figure*}

\subsection{Experimental Settings}
We validate our tokenization method by applying it to ViT, referring to the resulting model as \ours. As baselines, we include DeiT~\cite{deit} and three tokenization-focused approaches—SPiT\cite{spit}, Quadformer\cite{quadformer}, and MSViT\cite{msvit}—while keeping the backbone unchanged.

We evaluate three model scales: Tiny $(D=192)$, Small $(D=384)$, and Base $(D=768)$, setting the local and positional feature dimensions to $D/4$. All models process $224\times224$ images during training and evaluation.

For pixel-level embedding, we use a stride-2 convolution, reducing feature map size by half. The positional feature map $\mathbf{F}_{pos}$ and superpixel index map $\mathcal{I}$ are down-scaled accordingly. A preliminary study showed that using stride 1 yielded negligible gains at a much higher computational cost, so we opted for stride 2.

Among the baseline methods, SPiT~\cite{spit} is the most closely related to ours, as it also employs superpixels in tokenization. Meanwhile, Quadformer~\cite{quadformer} and MSViT~\cite{msvit} do not utilize superpixels but improve the tokenization process through mixed-scale tokenization. Following the settings of each method, we conduct experiments in two different settings.
For DeiT~\cite{deit} and SPiT~\cite{spit}, we train and evaluate the models using the standard ImageNet-1K~(IN-1K)~\cite{imagenet} classification setting~\cite{deit, resnet, swin, pvt, t2tvit}, where all models are trained from scratch. In contrast, for Quadformer~\cite{quadformer} and MSViT~\cite{msvit}, we follow their experimental setups. Specifically, MSViT initializes the transformer backbone with a ViT model pre-trained on ImageNet-21K~(IN-21K)~\cite{imagenet_21k}, while Quadformer uses a ViT model that is first pre-trained on IN-21K and then further fine-tuned on IN-1K. In our experiments, we follow the setting of Quadformer.

For the superpixel algorithm, we employ SLIC~\cite{slic}, but in practice, we use its faster implementation, FastSLIC~\cite{fastslic} due to its computational efficiency. We configure SLIC~\cite{slic} with its default setting, a compactness of 10, a maximum of 10 iterations, and 196 superpixels. Note that the value 196 is simply the maximum number of superpixels, and the actual number of superpixels varies across images. For example, on the ImageNet-1K~\cite{imagenet} validation set, the average number of superpixels is approximately 191.9 ± 6.0.

\subsection{Comparison to Baselines}
Table~\ref{tab:main_table} compares the classification performance of \ours with baselines in two different settings. For each model size, the table provides a detailed comparison across key aspects: number of total parameters of the model (\# Params.), computational cost measured in GMACs (GMACs), and the classification accuracy (Top-1).

The results in Table~\ref{tab:main_table} demonstrate that \ours consistently outperforms the baselines in both experimental settings. 
Specifically, when trained from scratch with randomly initialized weights, \ours achieves performance gains across all model sizes compared to DeiT~\cite{deit}: Tiny (+3.5\%p), Small (+1.1\%p), and Base (+0.5\%p). Additionally, \ours surpasses SPiT~\cite{spit}, the most closely related superpixel tokenization baseline, by substantial margins: Small (+5.9\%p) and Base (+1.9\%p).

In the setting where weights are initialized with IN-21K~\cite{imagenet_21k}, \ours consistently outperforms Quadformer~\cite{quadformer} and MSViT~\cite{msvit} across all model scales. Notably, in the Tiny model, \ours with random initialization achieves superior performance compared to MSViT~\cite{msvit} initialized with IN-21K~\cite{imagenet_21k} pre-trained weights. Furthermore, in the Small model, our approach trained with random initialization even surpasses Quadformer~\cite{quadformer} initialized with IN-21K~\cite{imagenet_21k} + IN-1K~\cite{imagenet} pre-trained weights.

\subsection{Adaptive Inference}
Figure \ref{fig:adaptive_inference} highlights the adaptive inference capability of \ours. This figure illustrates the performance variation with respect to the number of tokens during inference, where SPiT~\cite{spit} is plotted based on the average token count, while \ours is plotted using the maximum token count.
Unlike DeiT~\cite{deit}, which relies on a fixed number of tokens, \ours can dynamically adjust the number of superpixel tokens during inference, enabling more efficient computation while maintaining strong performance. Notably, \ours outperforms DeiT~\cite{deit} even with fewer tokens, requiring only 100 tokens compared to DeiT's 196 to achieve the same performance in the tiny model setting. This highlights its ability to extract meaningful visual representations with higher efficiency.

Additionally, \ours can further increase the number of superpixels at inference time, effectively scaling accuracy as needed. 
Notably, despite being trained with 196 superpixels, increasing to 256 at inference improves accuracy across different model scales: from 75.7\% to 75.9\% in the Tiny model, 80.9\% to 81.2\% in the Small model, and 82.3\% to 82.5\% in the Base model.
This contrasts with SPiT~\cite{spit}, which, despite its ability to adjust token counts, fails to deliver competitive performance across all token settings. In specific, SPiT-B~\cite{spit} achieves only 64.0\% accuracy with an average of 46 tokens, while increasing the token count to an average of 774 marginally improves performance to 79.2\%. 
Note that SPiT~\cite{spit} does not allow direct control over the number of tokens but instead relies on a ``level'' hyperparameter, limiting fine-grained adjustments and restricting evaluations to extreme settings (46 and 774 tokens).

These findings underscore the strength of \ours's adaptive inference, allowing it to flexibly balance computational cost and accuracy, making it a more scalable and efficient alternative to both DeiT~\cite{deit} and SPiT~\cite{spit}.

\subsection{Transfer Learning}
We verify the generalization ability of \ours when transferred to various downstream tasks. We follow the standard fine-tuning protocol and default settings of DeiT~\cite{deit} for fair comparison, by fine-tuning the ImageNet-1K pre-trained \ours to each downstream task specifically. We test our models on iNaturalist~\cite{inat18}, Flowers102~\cite{flowers}, and StanfordCars~\cite{cars}. Further details can be found in the Appendix~\ref{sec:appendix_transfer}.

\begin{table}[t!]
    \centering
    \resizebox{0.47\textwidth}{!}{%
    \begin{tabular}{c c c c c c}
    \toprule
         Size & Model & $\text{INat}_{18}$ & $\text{INat}_{19}$ & 
         Cars & Flowers\textsuperscript{\textdagger} \\
    \midrule
    \midrule
        \multirow{2}{*}{Small} & DeiT & 70.7 & 76.6 & \textbf{92.1} & 96.5 \\
        & \ours & \textbf{71.4} & \textbf{77.9} & \textbf{92.1} &\textbf{96.8}\\
    \midrule
        \multirow{2}{*}{Base} & DeiT & 73.2 & 77.7 & \textbf{92.1} & 96.8 \\
        & \ours & \textbf{74.0}  & \textbf{78.4} & \textbf{92.1} & \textbf{96.9} \\
    \bottomrule
    \end{tabular}
    }
    \caption{\textbf{Results of transfer learning different datasets in terms of top-1 accuracy results.} \ours outperforms ViT in most downstream tasks in both supervised and self-supervised settings. \textsuperscript{\textdagger}: The reported Flowers102 performance could not be reproduced in the exact setting, so we compare with our reproduced results.}
    \label{tab:transfer}
\end{table}

\begin{table*}[h!]
\centering
\resizebox{\textwidth}{!}{%
\begin{tabular}{lccccccccccc}
\toprule
 &    & \multicolumn{3}{c}{ECSSD} & \multicolumn{3}{c}{DUTS} & \multicolumn{3}{c}{DUT-OMRON} \\
\cmidrule{3-5} \cmidrule{6-8} \cmidrule{9-11}
Model  & Postproc.& IoU & Acc. & max $F_\beta$ & IoU & Acc. & max $F_\beta$ & IoU & Acc. & max $F_\beta$ \\
\midrule
\midrule
SPiT$^{\dagger}$~\cite{spit}& $\times$ & 
\underline{77.3} & \underline{93.4} &\textbf{90.3} & \textbf{63.9} & \underline{89.4} & \textbf{77.1} & \underline{56.4} & \underline{86.8} & \textbf{71.1} \\
\midrule
\textcolor{lightgray}{DINO-ViT~\cite{dino}} &
\textcolor{lightgray}{\checkmark} &
\textcolor{lightgray}{76.7} &
\textcolor{lightgray}{93.2} & 
\textcolor{lightgray}{86.9} & 
\textcolor{lightgray}{61.0} & 
\textcolor{lightgray}{90.6} & 
\textcolor{lightgray}{74.1} & 
\textcolor{lightgray}{59.9} & 
\textcolor{lightgray}{88.5} & 
\textcolor{lightgray}{67.6} \\
DINO-ViT~\cite{dino} & $\times$  & 
71.1 & 91.6 & 80.4 & 56.7 &\textbf{89.5} & 66.4 & 51.8 & \underline{86.8} & 58.5 \\
\midrule
\textcolor{lightgray}{DINO-\ours} & \textcolor{lightgray}{\checkmark}&
\textcolor{lightgray}{79.7} & 
\textcolor{lightgray}{93.8} & 
\textcolor{lightgray}{88.7} & 
\textcolor{lightgray}{60.2} & 
\textcolor{lightgray}{89.1} & 
\textcolor{lightgray}{71.3} & 
\textcolor{lightgray}{60.5} & 
\textcolor{lightgray}{87.9} & 
\textcolor{lightgray}{67.6} \\
DINO-\ours  & $\times$  & 
\textbf{80.5 (\textcolor{OliveGreen}{$\uparrow$ 9.4})} &
\textbf{93.8 (\textcolor{OliveGreen}{$\uparrow$ 2.2})} & 
\underline{87.0 (\textcolor{OliveGreen}{$\uparrow$ 6.6})} &
\underline{60.0 (\textcolor{OliveGreen}{$\uparrow$ 3.3})} &
88.8 (\textcolor{NavyBlue}{$\downarrow$ 0.7}) & 
\underline{68.0 (\textcolor{OliveGreen}{$\uparrow$ 1.6})} &
\textbf{57.5 (\textcolor{OliveGreen}{$\uparrow$ 5.7})} & 
\textbf{87.2 (\textcolor{OliveGreen}{$\uparrow$ 0.4})} & 
\underline{63.4 (\textcolor{OliveGreen}{$\uparrow$ 4.9})} \\
\bottomrule
\end{tabular}
}
\vspace{-0.5em}
\caption{\textbf{Results of zero-shot salient segmentation with TokenCut~\cite{tokencut}}. Models with additional postprocessing are colored in \textcolor{gray}{gray}. The best performance in each column is highlighted in bold, and the second-best performance is indicated with an underline among non-postprocessed results. $^{\dagger}$ denotes that the model is trained at a different training setting to others, \textit{i.e.,} with different resolution ($256$ vs $224$), data augmentation techniques and learning method (supervised vs self-supervised).}
\vspace{-1.5em}
\label{tab:sel_zero_seg}

\end{table*}

Table~\ref{tab:transfer} presents a comparison on the classification accuracy between DeiT and \ours when fine-tuned on various downstream tasks.
Across various tasks and model scales, \ours consistently achieves higher accuracy than DeiT. Notably, iNaturalist2018~\cite{inat}, a dataset of sufficient scale, SuiT-Small outperforms DeiT-Small by 0.8\%p, highlighting its superior ability to transfer learned features effectively especially in data-rich scenarios.

These results indicate that \ours shows improved generalizability than DeiT. This positions \ours a strong choice for transfer learning across diverse domains. The consistently superior performance of \ours, regardless of model size or dataset complexity, further underscores its suitability for real-world applications.

\subsection{Zero-shot Salient Object Segmentation}
We conduct zero-shot salient object segmentation to demonstrate the advantages of superpixel tokens in dense prediction tasks.
We adopt DINO~\cite{dino} models for this experiment.
We evaluate the performance on ECSSD~\cite{ecssd}, DUTS~\cite{duts} and DUT-OMRON~\cite{dut} using the TokenCut framework~\cite{tokencut}.

Superpixel tokenization significantly boosts DINO-ViT’s performance across most metrics, even without post-processing, as shown in Table~\ref{tab:sel_zero_seg}. Unlike DINO-ViT, \ours maintains minimal performance differences between processed and unprocessed results. Compared to SPiT~\cite{spit}, \ours achieves competitive performance, though SPiT was trained on higher-resolution images. 

\begin{figure}[!t]
    \centering
    \includegraphics[width=\linewidth]{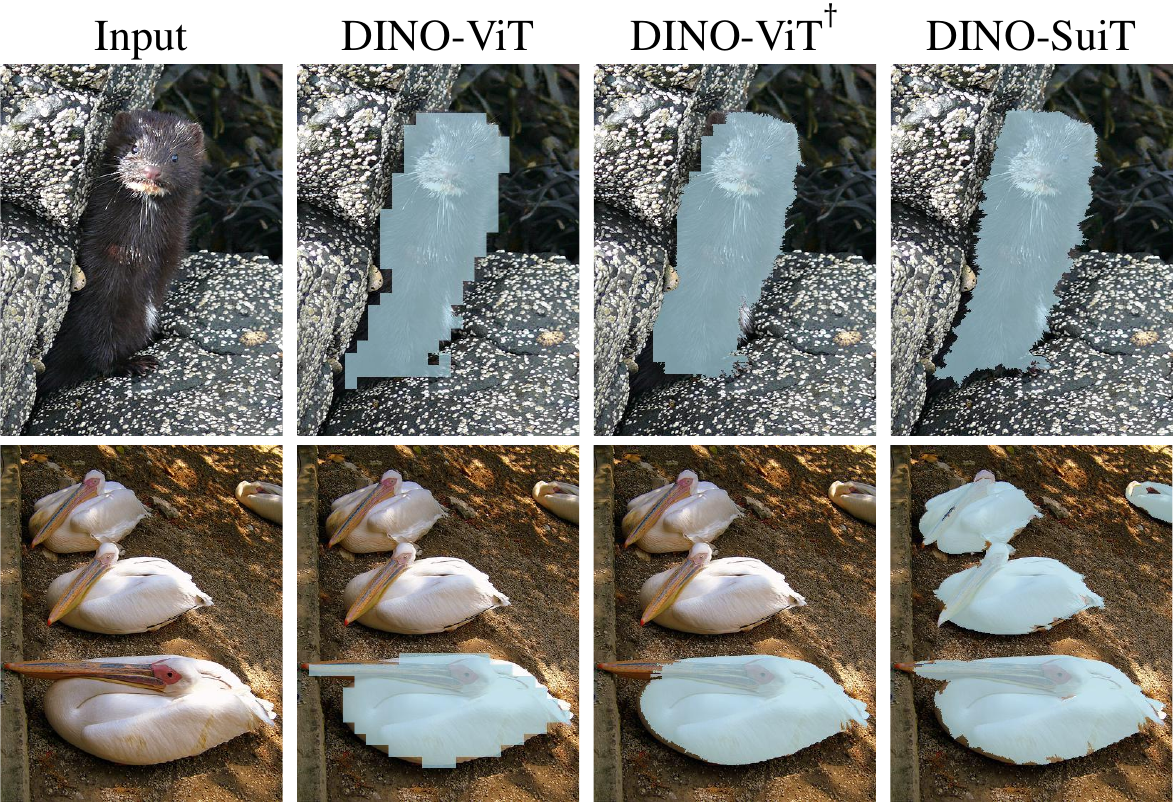} 
    \caption{\textbf{Qualitative results of zero-shot salient object segmentation}. $^{\dagger}$ denotes the model with additional post-processing. DINO-\ours successfully detects salient objects both in single- and multi-object scenarios without any post-processing.}
    \label{fig:zero_seg}
    \vspace{-1.5em}
\end{figure}

Moreover, as Figure~\ref{fig:zero_seg} shows, DINO-\ours detects all salient objects, even in multi-object scenarios, whereas DINO-ViT struggles with rectangular masks and incomplete detections. 
The visualization of self-attention maps of DINO-\ours can be found in the Appendix~\ref{appendix:selsup}.

\subsection{Semantic Integrity of Superpixel Tokens} 
To assess the semantic preservation of our superpixel tokens, we perform K-means clustering~\cite{kmeans} in the token embedding space and compare the results with those of patch tokens.

\begin{figure}[!t]
    \centering
    \includegraphics[width=\linewidth]{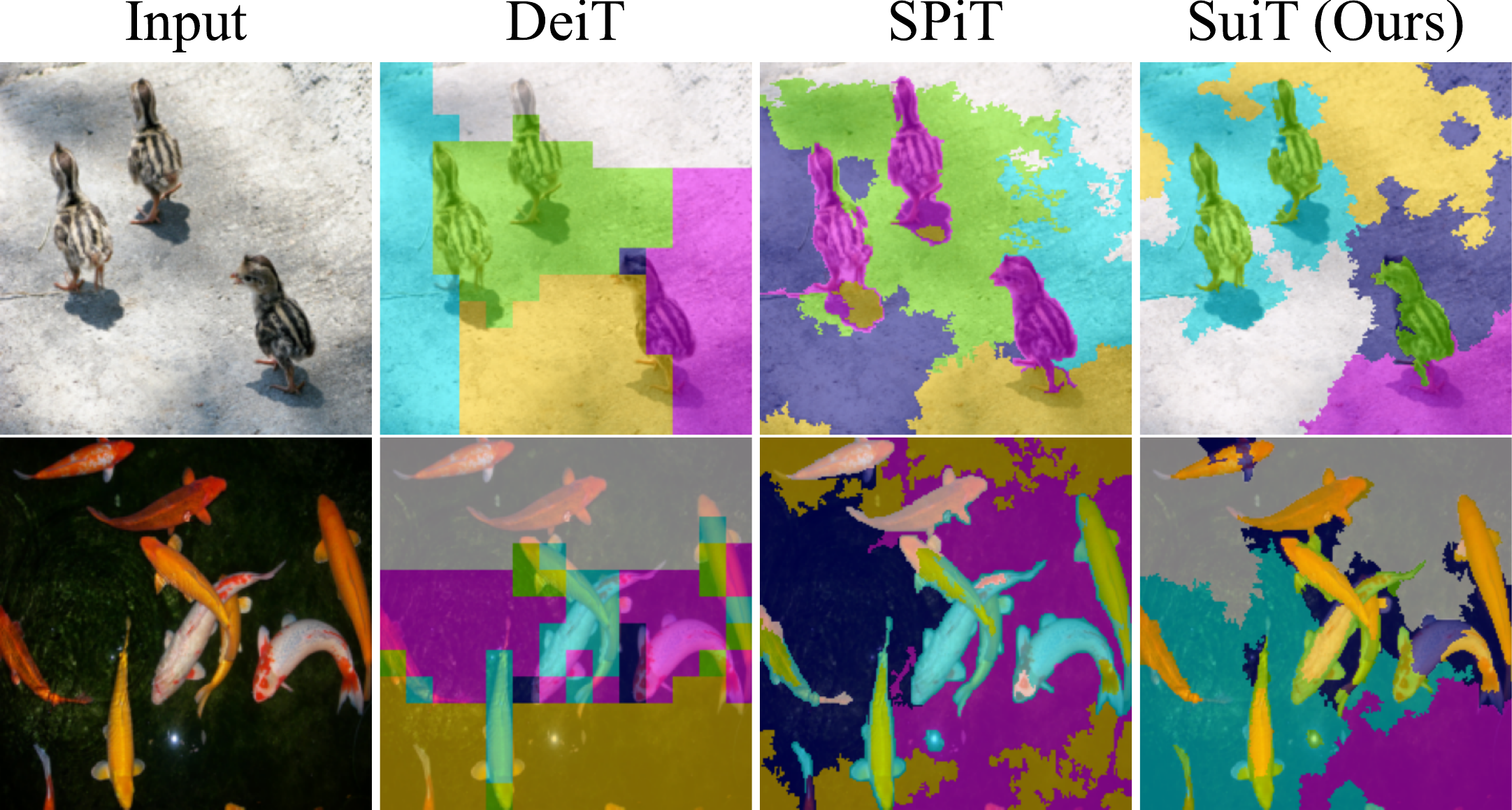} 
    \caption{\textbf{Comparison of K-means clustering results for patch tokens from DeiT and superpixel tokens from \ours, both supervisedly trained on ImageNet-1K, with K=6.} Superpixel tokens in \ours\ tend to cluster by semantic meaning, while ViT patch tokens mostly cluster by positional information.}
    \vspace{-1em}
    \label{fig:token_analysis} 
\end{figure}

As shown in Figure~\ref{fig:token_analysis}, our superpixel tokens cluster based on semantic similarity, whereas DeiT’s patch tokens lack consistency. For example, in the first row, tokens from duckling regions are grouped together, while DeiT’s patch tokens fail to capture clear semantic meaning. This suggests that superpixel tokens better preserve semantic integrity.

SPiT, which also utilizes superpixels in tokenization, exhibits similar behavior to \ours. However, SPiT clusters the shadows together with the ducklings in the first row, whereas those of \ours selectively groups only the ducklings. This further supports that our tokenization method preserves stronger semantic coherence.
Additional results can be found in the Appendix~\ref{sec:appendix_semantic_integrity}.

\begin{table}[t!]
    \centering
    \begin{tabular}{cccc}
    \toprule
       Local feats. & PE Freqs. & PE Injection & Top-1 \\
    \midrule\midrule
    \cmark & random init. & concat+proj & 75.7 \\
    \midrule
    \xmark & \textcolor{lightgray}{random init.} & \textcolor{lightgray}{concat+proj} &  66.5 \\
    \textcolor{lightgray}{\cmark} & \multicolumn{2}{c}{No PE} & 71.5 \\
    \textcolor{lightgray}{\cmark} & pre-defined & \textcolor{lightgray}{concat+proj} & 73.0 \\
    \textcolor{lightgray}{\cmark} & \textcolor{lightgray}{random init.} & concat & 73.8 \\
    \textcolor{lightgray}{\cmark} & \textcolor{lightgray}{random init.} & addition & 75.2 \\
    \bottomrule
    \end{tabular}
    \caption{\textbf{Ablation study on technical components of pre-aggregate feature extraction}.}
    \label{tab:ablation_feats}
\end{table}

\begin{table}[t!]
    \centering
    \begin{tabular}{ccccc}
    \toprule
        Avg & Std & Softmax & Max & Top-1\\
    \midrule\midrule
        \cmark & \textcolor{lightgray}{\xmark} & \textcolor{lightgray}{\xmark} & \cmark & 75.7 \\
    \midrule
        \cmark & \textcolor{lightgray}{\xmark} & \textcolor{lightgray}{\xmark} & \textcolor{lightgray}{\xmark} & 73.2\\
        \textcolor{lightgray}{\xmark} & \textcolor{lightgray}{\xmark} & \textcolor{lightgray}{\xmark} & \cmark & 75.5 \\
        \cmark & \cmark & \textcolor{lightgray}{\xmark} & \textcolor{lightgray}{\xmark} & Fail\\
        \cmark & \textcolor{lightgray}{\xmark} & \cmark & \textcolor{lightgray}{\xmark} & 75.5\\
    \bottomrule
    \end{tabular}
    \caption{\textbf{Ablation on our aggregation method}. `Fail' indicates that the corresponding approach was unsuccessful in training.}
    \label{tab:ablation_agg}
\end{table}

\begin{table}[h!]
    \centering
    \begin{tabular}{cccc}
    \toprule
         Compactness  & Iters & Algorithm & Top-1 \\
         \midrule\midrule
         10 & 10 & Fast-SLIC & 75.7 \\
         \midrule
         1 &  \textcolor{lightgray}{10} & \textcolor{lightgray}{Fast-SLIC} & 74.4 \\
         100 & \textcolor{lightgray}{10} & \textcolor{lightgray}{Fast-SLIC} & 75.5 \\
         Square & \textcolor{lightgray}{N/A} & \textcolor{lightgray}{N/A} & 75.1 \\
         \textcolor{lightgray}{10} & 1 & \textcolor{lightgray}{Fast-SLIC} & 75.4 \\
         \textcolor{lightgray}{10} &  20 & \textcolor{lightgray}{Fast-SLIC} & 75.6 \\
         \textcolor{lightgray}{0.001} &  \textcolor{lightgray}{N/A} & Watershed & 73.9 \\

         \midrule
    \bottomrule
    \end{tabular}
    \caption{\textbf{Ablation on the superpixel algorithm.}}
    \vspace{-1em}
    \label{tab:ablation_slic}
\end{table}
\subsection{Ablation Studies}

\paragraph{Pixel-level Embeddings}
The following ablation studies on our proposed components are conducted using \ours-Tiny.
Table~\ref{tab:ablation_feats} highlights the importance of components in obtaining pixel-level embeddings. First, using raw 3-channel RGB features (row 2) leads to a significant performance drop, confirming the need for higher-dimensional local features. Second, removing positional information (row 3) reduces performance but retains partial class prediction through local cues. Experiments also reveal that pre-defined positional frequencies (row 4) under-perform compared to learned frequencies, emphasizing the need for vision-specific customization. Lastly, our results show that the combination of concatenation and projection more effectively integrate local and positional features than simple concatenation (row 5) or addition (row 6) used in prior studies~\cite{transformer, vit, deit}.

\paragraph{Aggregation}
Table~\ref{tab:ablation_agg} presents the analysis of our aggregation method, which employs cardinality-independent operations. Removing max pooling (row 2) results in a notable performance drop, and removing average pooling (row 3) causes a slight decrease. We also tested standard deviation as an alternative to max pooling, but it proved numerically unstable and unsuitable for training (row 4). Softmax pooling (row 5), which computes a weighted sum of values within each superpixel, showed promising results but slightly underperformed compared to max pooling. This may be due to the softmax operation suppressing prominent features crucial for recognition, leading to less discriminative representations.

\paragraph{Superpixel Method}
In Table~\ref{tab:ablation_slic}, we present a study regarding the superpixel methodology. 
First, we examine the compactness parameter in SLIC, which controls the trade-off between spatial proximity and pixel content in superpixel formation (rows 2-4). Higher values produce more regular shapes but reduce semantic relevance. We compare compactness values of 1 (row 2) and 100 (row 3), along with square-shaped tokens (row 4) as an upper-bound of high compactness. Keeping other components unchanged, we find that increasing compactness degrades performance, supporting the importance of semantic integrity. However, an extremely low value (\textit{e.g.}, 1) also harms performance, likely due to insufficient spatial coherence.
Our framework relies on SLIC, which iteratively clusters pixels. We try different number of iterations in clustering as shown in Table~\ref{tab:ablation_slic}, but the number of iterations had little impact on performance (rows 5 and 6). We also test Watershed~\cite{watershed} (row 7), which results in a slight performance drop but remains robust.
\section{Limitation and Future Work}
Our method, in its current form, is limited to the vanilla ViT architecture\cite{vit}. Yet, since the ViT architecture is the most prevalent choice for vision architecture in many applications, \textit{e.g.}, CLIP~\cite{clip} and LLaVA~\cite{llava} in multi-modal learning, we believe that building an improvement for the ViT architecture is valuable. Applying superpixel tokenizations to other ViT variants, \textit{e.g.}, Swin Transformer~\cite{swin} remains a topic for future study.
\section{Conclusion}
In this work, we proposed a novel superpixel-based tokenization method for ViTs as an attempt to improve the visual tokenization process.
By leveraging superpixels, the proposed tokenization obtains tokens preserving semantic integrity, overcoming the issue of mixed semantics in grid-based patches. 
To incorporate superpixels that are irregular in shape, size, and location, we proposed a two-stage pipeline consisting of pixel-level embedding and superpixel-aware aggregation. 
Extensive experiments demonstrated the effectiveness and intriguing properties of our method.
We hope our promising results inspire further exploration of token design in ViTs, fostering advancements in both model performance and interpretability.

{
    \small
    \bibliographystyle{ieeenat_fullname}
    \bibliography{main}

\begin{thebibliography}{75}
\providecommand{\natexlab}[1]{#1}
\providecommand{\url}[1]{\texttt{#1}}
\expandafter\ifx\csname urlstyle\endcsname\relax
  \providecommand{\doi}[1]{doi: #1}\else
  \providecommand{\doi}{doi: \begingroup \urlstyle{rm}\Url}\fi

\bibitem[Aasan et~al.(2024)Aasan, Kolbj\o{}rnsen, Schistad~Solberg, and Ram\'irez~Rivera]{spit}
Marius Aasan, Odd Kolbj\o{}rnsen, Anne Schistad~Solberg, and Ad\'in Ram\'irez~Rivera.
\newblock A spitting image: Modular superpixel tokenization in vision transformers.
\newblock In \emph{{CVF/ECCV} More Exploration, Less Exploitation Workshop ({MELEX} {ECCVW})}, 2024.

\bibitem[Achanta et~al.(2012)Achanta, Shaji, Smith, Lucchi, Fua, and Süsstrunk]{slic}
Radhakrishna Achanta, Appu Shaji, Kevin Smith, Aurelien Lucchi, Pascal Fua, and Sabine Süsstrunk.
\newblock Slic superpixels compared to state-of-the-art superpixel methods.
\newblock \emph{IEEE Transactions on Pattern Analysis and Machine Intelligence}, 34\penalty0 (11):\penalty0 2274--2282, 2012.

\bibitem[Agarap(2018)]{relu}
Abien~Fred Agarap.
\newblock Deep learning using rectified linear units (relu).
\newblock \emph{CoRR}, abs/1803.08375, 2018.

\bibitem[Alchan(2020)]{fastslic}
Kim Alchan.
\newblock fast-slic, 2020.

\bibitem[Bapna et~al.(2018)Bapna, Chen, Firat, Cao, and Wu]{bapna2018training}
Ankur Bapna, Mia~Xu Chen, Orhan Firat, Yuan Cao, and Yonghui Wu.
\newblock Training deeper neural machine translation models with transparent attention.
\newblock \emph{arXiv preprint arXiv:1808.07561}, 2018.

\bibitem[Belrose et~al.(2023)Belrose, Furman, Smith, Halawi, Ostrovsky, McKinney, Biderman, and Steinhardt]{belrose2023eliciting}
Nora Belrose, Zach Furman, Logan Smith, Danny Halawi, Igor Ostrovsky, Lev McKinney, Stella Biderman, and Jacob Steinhardt.
\newblock Eliciting latent predictions from transformers with the tuned lens.
\newblock \emph{arXiv preprint arXiv:2303.08112}, 2023.

\bibitem[Brown et~al.(2020)Brown, Mann, Ryder, Subbiah, Kaplan, Dhariwal, Neelakantan, Shyam, Sastry, Askell, et~al.]{gpt3}
Tom~B Brown, Benjamin Mann, Nick Ryder, Melanie Subbiah, Jared Kaplan, Prafulla Dhariwal, Arvind Neelakantan, Pranav Shyam, Girish Sastry, Amanda Askell, et~al.
\newblock Language models are few-shot learners.
\newblock In \emph{Advances in Neural Information Processing Systems (NeurIPS)}, pages 1877--1901, 2020.

\bibitem[Carion et~al.(2020)Carion, Massa, Synnaeve, Usunier, Kirillov, and Zagoruyko]{detr}
Nicolas Carion, Francisco Massa, Gabriel Synnaeve, Nicolas Usunier, Alexander Kirillov, and Sergey Zagoruyko.
\newblock End-to-end object detection with transformers.
\newblock In \emph{European Conference on Computer Vision (ECCV)}, pages 213--229. Springer, 2020.

\bibitem[Caron et~al.(2021)Caron, Touvron, Misra, Jégou, Mairal, Bojanowski, and Joulin]{dino}
Mathilde Caron, Hugo Touvron, Ishan Misra, Hervé Jégou, Julien Mairal, Piotr Bojanowski, and Armand Joulin.
\newblock Emerging properties in self-supervised vision transformers, 2021.

\bibitem[Dehghan et~al.(2017)Dehghan, Masood, Shu, and Ortiz]{cars}
Afshin Dehghan, Syed~Zain Masood, Guang Shu, and Enrique~G. Ortiz.
\newblock View independent vehicle make, model and color recognition using convolutional neural network.
\newblock \emph{CoRR}, abs/1702.01721, 2017.

\bibitem[den Bergh et~al.(2013)den Bergh, Boix, Roig, and Gool]{seeds}
Michael~Van den Bergh, Xavier Boix, Gemma Roig, and Luc~Van Gool.
\newblock Seeds: Superpixels extracted via energy-driven sampling, 2013.

\bibitem[Deng et~al.(2009)Deng, Dong, Socher, Li, Li, and Fei-Fei]{imagenet}
Jia Deng, Wei Dong, Richard Socher, Li-Jia Li, Kai Li, and Li Fei-Fei.
\newblock Imagenet: A large-scale hierarchical image database.
\newblock In \emph{2009 IEEE conference on computer vision and pattern recognition}, pages 248--255. Ieee, 2009.

\bibitem[Djolonga et~al.(2021)Djolonga, Yung, Tschannen, Romijnders, Beyer, Kolesnikov, Puigcerver, Minderer, D'Amour, Moldovan, et~al.]{imageneta}
Josip Djolonga, Jessica Yung, Michael Tschannen, Rob Romijnders, Lucas Beyer, Alexander Kolesnikov, Joan Puigcerver, Matthias Minderer, Alexander D'Amour, Dan Moldovan, et~al.
\newblock On robustness and transferability of convolutional neural networks.
\newblock In \emph{Proceedings of the IEEE/CVF Conference on Computer Vision and Pattern Recognition}, pages 16458--16468, 2021.

\bibitem[Dosovitskiy et~al.(2021)Dosovitskiy, Beyer, Kolesnikov, Weissenborn, Zhai, Unterthiner, Dehghani, Minderer, Heigold, Gelly, Uszkoreit, and Houlsby]{vit}
Alexey Dosovitskiy, Lucas Beyer, Alexander Kolesnikov, Dirk Weissenborn, Xiaohua Zhai, Thomas Unterthiner, Mostafa Dehghani, Matthias Minderer, Georg Heigold, Sylvain Gelly, Jakob Uszkoreit, and Neil Houlsby.
\newblock An image is worth 16x16 words: Transformers for image recognition at scale.
\newblock In \emph{International Conference on Learning Representations}, 2021.

\bibitem[Esser et~al.(2021)Esser, Rombach, and Ommer]{vqgan}
Patrick Esser, Robin Rombach, and Bjorn Ommer.
\newblock Taming transformers for high-resolution image synthesis.
\newblock In \emph{Proceedings of the IEEE/CVF conference on computer vision and pattern recognition}, pages 12873--12883, 2021.

\bibitem[et~al.(2019)]{ace}
Ghorbani et al.
\newblock Towards automatic concept-based explanations.
\newblock In \emph{NeurIPS}, 2019.

\bibitem[et~al.(2023)]{eac}
Sun et al.
\newblock Explain any concept: Segment anything meets concept-based explanation.
\newblock In \emph{NeurIPS}, 2023.

\bibitem[Gao et~al.(2022)Gao, Li, Yang, Cheng, Han, and Torr]{gao2022large}
Shanghua Gao, Zhong-Yu Li, Ming-Hsuan Yang, Ming-Ming Cheng, Junwei Han, and Philip Torr.
\newblock Large-scale unsupervised semantic segmentation.
\newblock \emph{IEEE transactions on pattern analysis and machine intelligence}, 45\penalty0 (6):\penalty0 7457--7476, 2022.

\bibitem[Havtorn et~al.(2023)Havtorn, Royer, Blankevoort, and Bejnordi]{msvit}
Jakob~Drachmann Havtorn, Am{\'e}lie Royer, Tijmen Blankevoort, and Babak~Ehteshami Bejnordi.
\newblock Msvit: Dynamic mixed-scale tokenization for vision transformers.
\newblock In \emph{Proceedings of the IEEE/CVF International Conference on Computer Vision}, pages 838--848, 2023.

\bibitem[He et~al.(2016)He, Zhang, Ren, and Sun]{resnet}
Kaiming He, Xiangyu Zhang, Shaoqing Ren, and Jian Sun.
\newblock Deep residual learning for image recognition.
\newblock In \emph{Proceedings of the IEEE conference on computer vision and pattern recognition}, pages 770--778, 2016.

\bibitem[He et~al.(2015)He, Lau, Liu, Huang, and Yang]{he2015supercnn}
Shengfeng He, Rynson~WH Lau, Wenxi Liu, Zhe Huang, and Qingxiong Yang.
\newblock Supercnn: A superpixelwise convolutional neural network for salient object detection.
\newblock \emph{International journal of computer vision}, 115:\penalty0 330--344, 2015.

\bibitem[He et~al.(2018)He, Tan, Xia, He, Qin, Chen, and Liu]{he2018layer}
Tianyu He, Xu Tan, Yingce Xia, Di He, Tao Qin, Zhibo Chen, and Tie-Yan Liu.
\newblock Layer-wise coordination between encoder and decoder for neural machine translation.
\newblock \emph{Advances in Neural Information Processing Systems}, 31, 2018.

\bibitem[Hendrycks and Gimpel(2016)]{gelu}
Dan Hendrycks and Kevin Gimpel.
\newblock Gaussian error linear units (gelus).
\newblock \emph{arXiv preprint arXiv:1606.08415}, 2016.

\bibitem[Hendrycks et~al.(2021)Hendrycks, Zhao, Basart, Steinhardt, and Song]{imagenet_ao}
Dan Hendrycks, Kevin Zhao, Steven Basart, Jacob Steinhardt, and Dawn Song.
\newblock Natural adversarial examples.
\newblock In \emph{Proceedings of the IEEE/CVF Conference on Computer Vision and Pattern Recognition (CVPR)}, pages 15262--15271, 2021.

\bibitem[Horn et~al.(2017)Horn, Aodha, Song, Shepard, Adam, Perona, and Belongie]{inat}
Grant~Van Horn, Oisin~Mac Aodha, Yang Song, Alexander Shepard, Hartwig Adam, Pietro Perona, and Serge~J. Belongie.
\newblock The inaturalist challenge 2017 dataset.
\newblock \emph{CoRR}, abs/1707.06642, 2017.

\bibitem[Huang et~al.(2023)Huang, Zhou, Cao, He, and Tan]{supertoken}
Huaibo Huang, Xiaoqiang Zhou, Jie Cao, Ran He, and Tieniu Tan.
\newblock Vision transformer with super token sampling.
\newblock In \emph{Proceedings of the IEEE/CVF conference on computer vision and pattern recognition}, pages 22690--22699, 2023.

\bibitem[Ioffe(2015)]{batchnorm}
Sergey Ioffe.
\newblock Batch normalization: Accelerating deep network training by reducing internal covariate shift.
\newblock \emph{arXiv preprint arXiv:1502.03167}, 2015.

\bibitem[Jampani et~al.(2018)Jampani, Sun, Liu, Yang, and Kautz]{ssn}
Varun Jampani, Deqing Sun, Ming-Yu Liu, Ming-Hsuan Yang, and Jan Kautz.
\newblock Superpixel sampling networks.
\newblock In \emph{Proceedings of the European Conference on Computer Vision (ECCV)}, pages 352--368, 2018.

\bibitem[Jin and Han(2010)]{kmeans}
Xin Jin and Jiawei Han.
\newblock \emph{K-Means Clustering}, pages 563--564.
\newblock Springer US, Boston, MA, 2010.

\bibitem[Kenton and Toutanova(2019)]{kenton2019bert}
Jacob Devlin Ming-Wei~Chang Kenton and Lee~Kristina Toutanova.
\newblock Bert: Pre-training of deep bidirectional transformers for language understanding.
\newblock In \emph{Proceedings of naacL-HLT}, page~2. Minneapolis, Minnesota, 2019.

\bibitem[Kim et~al.(2024)Kim, Martino, and Sapiro]{svit}
Young~Kyung Kim, J.~Matías~Di Martino, and Guillermo Sapiro.
\newblock Vision transformers with natural language semantics, 2024.

\bibitem[Kudo(2018)]{kudo2018sentencepiece}
T Kudo.
\newblock Sentencepiece: A simple and language independent subword tokenizer and detokenizer for neural text processing.
\newblock \emph{arXiv preprint arXiv:1808.06226}, 2018.

\bibitem[Kwak et~al.(2017)Kwak, Hong, and Han]{kwak2017weakly}
Suha Kwak, Seunghoon Hong, and Bohyung Han.
\newblock Weakly supervised semantic segmentation using superpixel pooling network.
\newblock In \emph{Proceedings of the AAAI conference on artificial intelligence}, 2017.

\bibitem[Lewis(2019)]{lewis2019bart}
M Lewis.
\newblock Bart: Denoising sequence-to-sequence pre-training for natural language generation, translation, and comprehension.
\newblock \emph{arXiv preprint arXiv:1910.13461}, 2019.

\bibitem[Li et~al.(2024)Li, Ding, Yuan, Zhang, Pang, Cheng, Chen, Liu, and Loy]{li2024transformer}
Xiangtai Li, Henghui Ding, Haobo Yuan, Wenwei Zhang, Jiangmiao Pang, Guangliang Cheng, Kai Chen, Ziwei Liu, and Chen~Change Loy.
\newblock Transformer-based visual segmentation: A survey.
\newblock \emph{IEEE Transactions on Pattern Analysis and Machine Intelligence}, 2024.

\bibitem[Li and Chen(2015)]{lsc}
Zhengqin Li and Jiansheng Chen.
\newblock Superpixel segmentation using linear spectral clustering.
\newblock In \emph{Proceedings of the IEEE Conference on Computer Vision and Pattern Recognition (CVPR)}, 2015.

\bibitem[Liu et~al.(2023)Liu, Li, Wu, and Lee]{llava}
Haotian Liu, Chunyuan Li, Qingyang Wu, and Yong~Jae Lee.
\newblock Visual instruction tuning.
\newblock \emph{Advances in neural information processing systems}, 36:\penalty0 34892--34916, 2023.

\bibitem[Liu et~al.(2019)Liu, Ott, Goyal, Du, Joshi, Chen, Levy, Lewis, Zettlemoyer, and Stoyanov]{liu2019roberta}
Yinhan Liu, Myle Ott, Naman Goyal, Jingfei Du, Mandar Joshi, Danqi Chen, Omer Levy, Mike Lewis, Luke Zettlemoyer, and Veselin Stoyanov.
\newblock Roberta: A robustly optimized bert pretraining approach.
\newblock \emph{arXiv e-prints}, pages arXiv--1907, 2019.

\bibitem[Liu et~al.(2021)Liu, Lin, Cao, Hu, Wei, Zhang, Lin, and Guo]{swin}
Ze Liu, Yutong Lin, Yue Cao, Han Hu, Yixuan Wei, Zheng Zhang, Stephen Lin, and Baining Guo.
\newblock Swin transformer: Hierarchical vision transformer using shifted windows.
\newblock In \emph{Proceedings of the IEEE/CVF international conference on computer vision}, pages 10012--10022, 2021.

\bibitem[Ma et~al.()Ma, Zhou, Wang, Qin, Sun, Liu, and Fu]{coc}
Xu Ma, Yuqian Zhou, Huan Wang, Can Qin, Bin Sun, Chang Liu, and Yun Fu.
\newblock Image as set of points.
\newblock In \emph{The Eleventh International Conference on Learning Representations}.

\bibitem[Mei et~al.(2024)Mei, Chen, Yuille, and Xie]{spformer}
Jieru Mei, Liang-Chieh Chen, Alan Yuille, and Cihang Xie.
\newblock Spformer: Enhancing vision transformer with superpixel representation.
\newblock \emph{arXiv preprint arXiv:2401.02931}, 2024.

\bibitem[Mo et~al.(2023)Mo, Xie, Chu, Hong, Niessner, and Li]{dit}
Shentong Mo, Enze Xie, Ruihang Chu, Lanqing Hong, Matthias Niessner, and Zhenguo Li.
\newblock Dit-3d: Exploring plain diffusion transformers for 3d shape generation.
\newblock \emph{Advances in neural information processing systems}, 36:\penalty0 67960--67971, 2023.

\bibitem[Neubert and Protzel(2014)]{watershed}
Peer Neubert and Peter Protzel.
\newblock Compact watershed and preemptive slic: On improving trade-offs of superpixel segmentation algorithms.
\newblock In \emph{ICPR}, 2014.

\bibitem[Nilsback and Zisserman(2008)]{flowers}
Maria-Elena Nilsback and Andrew Zisserman.
\newblock Automated flower classification over a large number of classes.
\newblock In \emph{Indian Conference on Computer Vision, Graphics and Image Processing}, 2008.

\bibitem[nostalgebraist(2020)]{nostalgebraist2020logitlens}
nostalgebraist.
\newblock interpreting gpt: the logit lens.
\newblock \emph{LessWrong}, 2020.

\bibitem[Radford et~al.()Radford, Wu, Child, Luan, Amodei, and Sutskever]{radfordlanguage}
Alec Radford, Jeffrey Wu, Rewon Child, David Luan, Dario Amodei, and Ilya Sutskever.
\newblock Language models are unsupervised multitask learners.

\bibitem[Radford et~al.(2021)Radford, Kim, Hallacy, Ramesh, Goh, Agarwal, Sastry, Askell, Mishkin, Clark, et~al.]{clip}
Alec Radford, Jong~Wook Kim, Chris Hallacy, Aditya Ramesh, Gabriel Goh, Sandhini Agarwal, Saurabh Sastry, Amanda Askell, Pamela Mishkin, Jack Clark, et~al.
\newblock Learning transferable visual models from natural language supervision.
\newblock In \emph{Proceedings of the International Conference on Machine Learning (ICML)}, 2021.

\bibitem[Ridnik et~al.(2021)Ridnik, Baruch, Noy, and Zelnik{-}Manor]{imagenet_21k}
Tal Ridnik, Emanuel~Ben Baruch, Asaf Noy, and Lihi Zelnik{-}Manor.
\newblock Imagenet-21k pretraining for the masses.
\newblock \emph{CoRR}, abs/2104.10972, 2021.

\bibitem[Ronen et~al.(2023)Ronen, Levy, and Golbert]{quadformer}
Tomer Ronen, Omer Levy, and Avram Golbert.
\newblock Vision transformers with mixed-resolution tokenization.
\newblock In \emph{Proceedings of the IEEE/CVF Conference on Computer Vision and Pattern Recognition}, pages 4613--4622, 2023.

\bibitem[Shehzadi et~al.(2023)Shehzadi, Hashmi, Stricker, and Afzal]{shehzadi20232d}
Tahira Shehzadi, Khurram~Azeem Hashmi, Didier Stricker, and Muhammad~Zeshan Afzal.
\newblock 2d object detection with transformers: a review.
\newblock \emph{arXiv preprint arXiv:2306.04670}, 2023.

\bibitem[Shibata et~al.(1999)Shibata, Kida, Fukamachi, Takeda, Shinohara, Shinohara, and Arikawa]{bytepairencoding}
Yusuxke Shibata, Takuya Kida, Shuichi Fukamachi, Masayuki Takeda, Ayumi Shinohara, Takeshi Shinohara, and Setsuo Arikawa.
\newblock Byte pair encoding: A text compression scheme that accelerates pattern matching.
\newblock 1999.

\bibitem[Song et~al.(2021)Song, Salcianu, Song, Dopson, and Zhou]{wordpiece}
Xinying Song, Alex Salcianu, Yang Song, Dave Dopson, and Denny Zhou.
\newblock Fast wordpiece tokenization.
\newblock In \emph{Proceedings of the 2021 Conference on Empirical Methods in Natural Language Processing}, pages 2089--2103, 2021.

\bibitem[Strudel et~al.(2021)Strudel, Garcia, Laptev, and Schmid]{segmenter}
Robin Strudel, Ricardo Garcia, Ivan Laptev, and Cordelia Schmid.
\newblock Segmenter: Transformer for semantic segmentation.
\newblock In \emph{Proceedings of the IEEE/CVF international conference on computer vision}, pages 7262--7272, 2021.

\bibitem[Tancik et~al.(2020)Tancik, Srinivasan, Mildenhall, Fridovich-Keil, Raghavan, Singhal, Ramamoorthi, Barron, and Ng]{fourierfeatures}
Matthew Tancik, Pratul Srinivasan, Ben Mildenhall, Sara Fridovich-Keil, Nithin Raghavan, Utkarsh Singhal, Ravi Ramamoorthi, Jonathan Barron, and Ren Ng.
\newblock Fourier features let networks learn high frequency functions in low dimensional domains.
\newblock \emph{Advances in neural information processing systems}, 33:\penalty0 7537--7547, 2020.

\bibitem[Touvron et~al.(2021)Touvron, Cord, Douze, Massa, Sablayrolles, and J{\'e}gou]{deit}
Hugo Touvron, Matthieu Cord, Matthijs Douze, Francisco Massa, Alexandre Sablayrolles, and Herv{\'e} J{\'e}gou.
\newblock Training data-efficient image transformers \& distillation through attention.
\newblock In \emph{International conference on machine learning}, pages 10347--10357. PMLR, 2021.

\bibitem[Touvron et~al.(2023)Touvron, Lavril, Izacard, Martinet, Lachaux, Lacroix, Rozi{\`e}re, Goyal, Hambro, Azhar, et~al.]{touvron2023llama}
Hugo Touvron, Thibaut Lavril, Gautier Izacard, Xavier Martinet, Marie-Anne Lachaux, Timoth{\'e}e Lacroix, Baptiste Rozi{\`e}re, Naman Goyal, Eric Hambro, Faisal Azhar, et~al.
\newblock Llama: Open and efficient foundation language models.
\newblock \emph{arXiv preprint arXiv:2302.13971}, 2023.

\bibitem[Van~Horn et~al.(2018)Van~Horn, Mac~Aodha, Song, Cui, Sun, Shepard, Adam, Perona, and Belongie]{inat18}
Grant Van~Horn, Oisin Mac~Aodha, Yang Song, Yin Cui, Chen Sun, Alex Shepard, Hartwig Adam, Pietro Perona, and Serge Belongie.
\newblock The inaturalist species classification and detection dataset.
\newblock In \emph{Proceedings of the IEEE Conference on Computer Vision and Pattern Recognition (CVPR)}, 2018.

\bibitem[Vaswani et~al.(2017)Vaswani, Shazeer, Parmar, Uszkoreit, Jones, Gomez, Kaiser, and Polosukhin]{transformer}
Ashish Vaswani, Noam Shazeer, Niki Parmar, Jakob Uszkoreit, Llion Jones, Aidan~N Gomez, \L~ukasz Kaiser, and Illia Polosukhin.
\newblock Attention is all you need.
\newblock In \emph{Advances in Neural Information Processing Systems}. Curran Associates, Inc., 2017.

\bibitem[Vilas et~al.(2024)Vilas, Schauml{\"o}ffel, and Roig]{vilas2024analyzing}
Martina~G Vilas, Timothy Schauml{\"o}ffel, and Gemma Roig.
\newblock Analyzing vision transformers for image classification in class embedding space.
\newblock In \emph{Advances in Neural Information Processing Systems (NeurIPS)}, 2024.

\bibitem[Walmer et~al.(2023)Walmer, Suri, Gupta, and Shrivastava]{walmer2023teaching}
Matthew Walmer, Saksham Suri, Kamal Gupta, and Abhinav Shrivastava.
\newblock Teaching matters: Investigating the role of supervision in vision transformers.
\newblock In \emph{Proceedings of the IEEE/CVF Conference on Computer Vision and Pattern Recognition}, pages 7486--7496, 2023.

\bibitem[Wang et~al.(2017)Wang, Lu, Wang, Feng, Wang, Yin, and Ruan]{duts}
Lijun Wang, Huchuan Lu, Yifan Wang, Mengyang Feng, Dong Wang, Baocai Yin, and Xiang Ruan.
\newblock Learning to detect salient objects with image-level supervision.
\newblock In \emph{Proceedings of the IEEE conference on computer vision and pattern recognition}, pages 136--145, 2017.

\bibitem[Wang et~al.(2019)Wang, Li, Xiao, Zhu, Li, Wong, and Chao]{wang2019learning}
Qiang Wang, Bei Li, Tong Xiao, Jingbo Zhu, Changliang Li, Derek~F Wong, and Lidia~S Chao.
\newblock Learning deep transformer models for machine translation.
\newblock In \emph{Proceedings of the 57th Annual Meeting of the Association for Computational Linguistics}, pages 1810--1822, 2019.

\bibitem[Wang et~al.(2021)Wang, Xie, Li, Fan, Song, Liang, Lu, Luo, and Shao]{pvt}
Wenhai Wang, Enze Xie, Xiang Li, Deng-Ping Fan, Kaitao Song, Ding Liang, Tong Lu, Ping Luo, and Ling Shao.
\newblock Pyramid vision transformer: A versatile backbone for dense prediction without convolutions.
\newblock In \emph{Proceedings of the IEEE/CVF international conference on computer vision}, pages 568--578, 2021.

\bibitem[Wang et~al.(2022)Wang, Shen, Hu, Yuan, Crowley, and Vaufreydaz]{tokencut}
Yangtao Wang, Xi Shen, Shell~Xu Hu, Yuan Yuan, James~L Crowley, and Dominique Vaufreydaz.
\newblock Self-supervised transformers for unsupervised object discovery using normalized cut.
\newblock In \emph{Proceedings of the IEEE/CVF Conference on Computer Vision and Pattern Recognition}, pages 14543--14553, 2022.

\bibitem[Yan et~al.(2013)Yan, Xu, Shi, and Jia]{ecssd}
Qiong Yan, Li Xu, Jianping Shi, and Jiaya Jia.
\newblock Hierarchical saliency detection.
\newblock In \emph{Proceedings of the IEEE conference on computer vision and pattern recognition}, pages 1155--1162, 2013.

\bibitem[Yang et~al.(2013)Yang, Zhang, Lu, Ruan, and Yang]{dut}
Chuan Yang, Lihe Zhang, Huchuan Lu, Xiang Ruan, and Ming-Hsuan Yang.
\newblock Saliency detection via graph-based manifold ranking.
\newblock In \emph{Proceedings of the IEEE conference on computer vision and pattern recognition}, pages 3166--3173, 2013.

\bibitem[Yang et~al.(2020)Yang, Sun, Jin, and Zhou]{supfcn}
Fengting Yang, Qian Sun, Hailin Jin, and Zihan Zhou.
\newblock Superpixel segmentation with fully convolutional networks.
\newblock In \emph{Proceedings of the IEEE/CVF conference on computer vision and pattern recognition}, pages 13964--13973, 2020.

\bibitem[Yuan et~al.(2021)Yuan, Chen, Wang, Yu, Shi, Jiang, Tay, Feng, and Yan]{t2tvit}
Li Yuan, Yunpeng Chen, Tao Wang, Weihao Yu, Yujun Shi, Zi-Hang Jiang, Francis~EH Tay, Jiashi Feng, and Shuicheng Yan.
\newblock Tokens-to-token vit: Training vision transformers from scratch on imagenet.
\newblock In \emph{Proceedings of the IEEE/CVF international conference on computer vision}, pages 558--567, 2021.

\bibitem[Zhang et~al.(2022{\natexlab{a}})Zhang, Gu, Zhang, Bao, Chen, Wen, Wang, and Guo]{zhang2022styleswin}
Bowen Zhang, Shuyang Gu, Bo Zhang, Jianmin Bao, Dong Chen, Fang Wen, Yong Wang, and Baining Guo.
\newblock Styleswin: Transformer-based gan for high-resolution image generation.
\newblock In \emph{Proceedings of the IEEE/CVF conference on computer vision and pattern recognition}, pages 11304--11314, 2022{\natexlab{a}}.

\bibitem[Zhang et~al.(2018)Zhang, Cisse, Dauphin, and Lopez-Paz]{mixup}
Hongyi Zhang, Moustapha Cisse, Yann~N. Dauphin, and David Lopez-Paz.
\newblock mixup: Beyond empirical risk minimization.
\newblock In \emph{International Conference on Learning Representations}, 2018.

\bibitem[Zhang et~al.(2022{\natexlab{b}})Zhang, Pang, and Lu]{proxy}
Yifan Zhang, Bo Pang, and Cewu Lu.
\newblock Semantic segmentation by early region proxy.
\newblock In \emph{Proceedings of the IEEE/CVF Conference on Computer Vision and Pattern Recognition}, pages 1258--1268, 2022{\natexlab{b}}.

\bibitem[Zheng et~al.(2021)Zheng, Lu, Zhao, Zhu, Luo, Wang, Fu, Feng, Xiang, Torr, et~al.]{setr}
Sixiao Zheng, Jiachen Lu, Hengshuang Zhao, Xiatian Zhu, Zekun Luo, Yabiao Wang, Yanwei Fu, Jianfeng Feng, Tao Xiang, Philip~HS Torr, et~al.
\newblock Rethinking semantic segmentation from a sequence-to-sequence perspective with transformers.
\newblock In \emph{Proceedings of the IEEE/CVF conference on computer vision and pattern recognition}, pages 6881--6890, 2021.

\bibitem[Zhong et~al.(2020)Zhong, Zheng, Kang, Li, and Yang]{randomerasing}
Zhun Zhong, Liang Zheng, Guoliang Kang, Shaozi Li, and Yi Yang.
\newblock Random erasing data augmentation.
\newblock In \emph{Proceedings of the AAAI conference on artificial intelligence}, pages 13001--13008, 2020.

\bibitem[Zhu et~al.(2020)Zhu, Su, Lu, Li, Wang, and Dai]{deform_detr}
Xizhou Zhu, Weijie Su, Lewei Lu, Bin Li, Xiaogang Wang, and Jifeng Dai.
\newblock Deformable detr: Deformable transformers for end-to-end object detection.
\newblock \emph{arXiv preprint arXiv:2010.04159}, 2020.

\bibitem[Zhuang et~al.(2020)Zhuang, Qi, Duan, Xi, Zhu, Zhu, Xiong, and He]{transferlearning}
Fuzhen Zhuang, Zhiyuan Qi, Keyu Duan, Dongbo Xi, Yongchun Zhu, Hengshu Zhu, Hui Xiong, and Qing He.
\newblock A comprehensive survey on transfer learning.
\newblock \emph{Proceedings of the IEEE}, 109\penalty0 (1):\penalty0 43--76, 2020.

\end{thebibliography}
}

\clearpage
\setcounter{table}{0}
\renewcommand{\thetable}{A\arabic{table}}
\setcounter{figure}{0}
\renewcommand{\thefigure}{A\arabic{figure}}
\setcounter{section}{0}
\renewcommand{\thesection}{A\arabic{section}}
\maketitlesupplementary

\definecolor{citecolor}{RGB}{60,130,200}

\begin{figure}[!t]
    \centering
    \includegraphics[width=\linewidth]{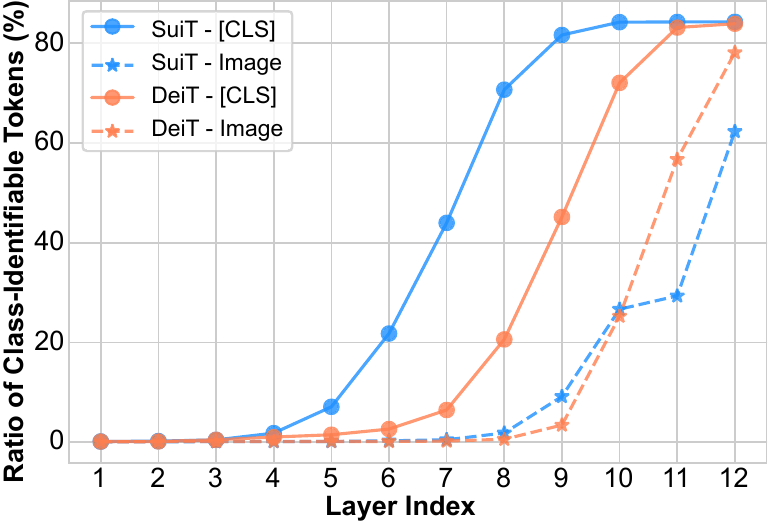}
    \caption{\textbf{Class-identifiability analysis results of DeiT and \ours.} We show the proportion of class-identifiable [CLS] tokens and image tokens for SuiT and DeiT across all layers.}
    \label{fig:class_ident} 
\end{figure}

\begin{table}[t!]
    \centering
    \begin{tabular}{ccccc}
    \toprule
         \multirow{2}{*}{Size} &  \multirow{2}{*}{Model} & ImageNet-A & ImageNet-O \\
         & & (Acc.) & (AUPR) \\
    \midrule\midrule
        \multirow{2}{*}{Tiny} & DeiT & 7.0 & 17.4 \\
        & \ours & \textbf{11.1} & \textbf{19.8} \\
    \midrule
        \multirow{2}{*}{Small} & DeiT & 19.2 & 20.9 \\
        & \ours & \textbf{22.5} & \textbf{23.3} \\
    \midrule
        \multirow{2}{*}{Base} & DeiT & 27.9 & 24.6 \\
        & \ours & \textbf{29.5}  & \textbf{26.7} \\
    \bottomrule
    \end{tabular}%
    \caption{\textbf{Experimental results of~\ours and DeiT on ImageNet-A and ImageNet-O}. We report accuracy on ImageNet-A and AUPR on ImageNet-O, following ~\cite{imagenet_ao}.}
    \label{tab:imagenet_variants}
\end{table}

\section{Additional Experiments}
\subsection{Class Identifiability Analysis}
\label{sec:appendix_clsid}
\cite{vilas2024analyzing} examined class-specific encoding in supervised ViTs by projecting intermediate representations through the trained classifier head into a class-specific embedding space, similar to the logit lens method in NLP~\cite{nostalgebraist2020logitlens, belrose2023eliciting}. Building on this, we evaluate class-specific encoding across SuiT’s layers by projecting token representations through the classifier head and measuring the proportion of tokens correctly predicting the target class on IN-S-919~\cite{gao2022large}.

Figure~\ref{fig:class_ident} shows that SuiT’s classification token shows higher class identifiability in the earlier layers compared to DeiT. This can be attributed to the semantic preserving nature of superpixel tokenization, which aids in aggregating class-specific encoding in early layers. In the deeper layers, DeiT’s patch tokens exhibit class identifiability levels slightly lower than its classification token, with a marginal gap. This observation aligns with findings from previous studies~\cite{vilas2024analyzing, walmer2023teaching}, which report that patch tokens in supervised ViTs tend to encode increasingly global information as layers deepen. In contrast, SuiT’s patch tokens maintain a lower proportion of class-identifiable features in the later layers, suggesting a reduced emphasis on global feature encoding. This divergence alludes that SuiT and DeiT adopt distinctive strategies to aggregate class-specific features at different network depths.

\subsection{Enhanced Robustness} We examine the robustness and out-of-distribution (OOD) generalization ability of \ours in ImageNet variants, ImageNet-A~\cite{imageneta} and ImageNet-O~\cite{imagenet_ao}. ImageNet-A is a subset of ImageNet-1K curated to mislead classifiers and test model robustness, while ImageNet-O, composed of OOD samples not in ImageNet-1K, evaluates the OOD detection capabilities of models. 

In Table~\ref{tab:imagenet_variants}, \ours consistently outperforms DeiT on both datasets across model variants.
These findings indicate that our superpixel tokenization improves model robustness in challenging scenarios by mitigating the model's reliance on features that are easily learned but lack generalizability.

\section{Implementation Details}

\subsection{ImageNet-1K classification}
\subsubsection{Training from scratch}
The settings for experiments from scratch is presented in Table~\ref{tab:in1k_settings_scratch}. We mostly follow the settings of DeiT~\cite{deit} with minimal changes.
The main difference to the DeiT setting is as follows:
1) to align with our main concept of using superpixels, we disable 
MixUp~\cite{mixup}.
2) In random erasing~\cite{randomerasing},
the randomly erased region is filled with per-channel random color instead of per-pixel random color. 3) We use gradient clipping, which was known to be helpful in ViT~\cite{vit}, but later removed in DeiT~\cite{deit}.
Other than that, we slightly tweak hyper-parameters of learning rate, weight decay and stochastic depth.
For models of all three scales, Tiny, Small and Base, we use a base learning rate of $8e-4$, which is scaled proportionally to the batch size, by $\times \frac{batchsize}{512}$. We use different weight decay and stochastic depth values for each scale. Most notably, our model uses a larger weight decay and drop rate for stochastic depth, especially in Base scale. This is because we found that our model tends to overfit quickly, compared to the baseline DeiT. We thus opt to use a stronger regularization term. Similar observations were made in Transfer Learning, which is also discussed in the Section ~\ref{sec:appendix_transfer}.

\subsubsection{Fine-tuning from Pre-trained weights}
The settings for fine-tuning is presented in Table~\ref{tab:in1k_settings_finetune}. We mostly follow the settings of Ronen et al.~\cite{quadformer}. We initialize our models with the weights pre-trained on ImageNet-21K and fine-tuned on Imagenet-1K, as in \cite{quadformer}, and the tokenization part is randomly initialized only. We use the pre-trained weights provided by the \texttt{timm} library, \texttt{vit\_\{size\}\_patch16\_224.augreg\_in21k\_ft\_in1k}, where \texttt{\{size\}} denotes the model size, \textit{e.g.}, Tiny, Small, Base. We train our models for 130 epochs. In this experiment, stronger weight decay did not lead to notable difference, but the drop rate of stochastic depth was significant in prevention of overfitting, as in the experiments from scratch.

\begin{figure}[t!]
    \centering
    \includegraphics[width=\linewidth]{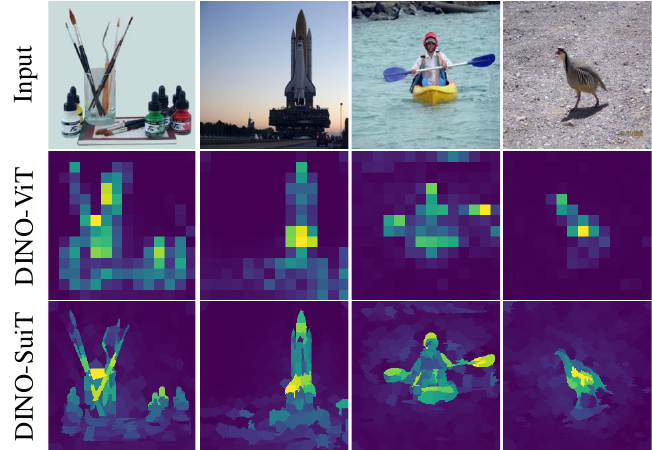}
    \caption{\textbf{Self-attention map comparison between DINO-ViT and DINO-\ours.} All maps are visualized by averaging the [CLS] token's self-attention from the final layer across all heads.}
    \label{fig:dino_attention}
\end{figure}

\subsection{Transfer Learning}
\label{sec:appendix_transfer}
Here, we list the training settings used for transfer learning. For both ImageNet-1K~\cite{imagenet} classification models and  DINO~\cite{dino} pre-trained models, we follow the default training settings of DeiT~\cite{deit} and DINO~\cite{dino}. The details for each experiment can be found at Table~\ref{tab:transfer_learning_appendix} and Table~\ref{tab:transfer_learning_dino_appendix}.
Our experiments, especially on smaller datasets, required stronger regularization terms. A very high drop rate for stochastic depth of 0.5 was essential in Flowers and StandfordCars, to avoid overfitting.

\begin{table*}[t!]
    \centering
    \begin{minipage}{0.4\textwidth}
        \centering
        \resizebox{\textwidth}{!}{
    \begin{tabular}{l ccc}
    \toprule
        Model size & Tiny & Small & Base \\
    \midrule\midrule
        Epochs & \multicolumn{3}{c}{300} \\
    \midrule
        Batch size & \multicolumn{3}{c}{1024} \\
        Optimizer & \multicolumn{3}{c}{AdamW} \\
        Base Learning rate & \multicolumn{3}{c}{8e-4} \\
        Learning rate decay & \multicolumn{3}{c}{cosine} \\
        Weight decay & 0.02 & 0.05 & 0.08 \\
        Warmup epochs & \multicolumn{3}{c}{5} \\
    \midrule
        Label smoothing & \multicolumn{3}{c}{0.1}\\
        Dropout & \multicolumn{3}{c}{\xmark} \\
        Stoch. Depth & 0.0 & 0.1 & 0.3 \\
        Repeated Aug & \multicolumn{3}{c}{\cmark} \\
        Gradient Clip. & \multicolumn{3}{c}{\cmark} \\
    \midrule
        Rand Augment & \multicolumn{3}{c}{9 / 0.5}\\
        Mixup prob. & \multicolumn{3}{c}{0.0} \\
        Cutmix prob. & \multicolumn{3}{c}{1.0} \\
        Erasing prob. & \multicolumn{3}{c}{0.25} \\ 
    \bottomrule
    \end{tabular}
    }
    \caption{\textbf{Settings of \ours on ImageNet-1K classification, training from scratch.} We mostly follow the settings open to public by the authors~\cite{deit}.}
    \label{tab:in1k_settings_scratch}
    \end{minipage}
    \hspace{1cm}
    \begin{minipage}{0.4\textwidth}
        \centering
        \resizebox{\textwidth}{!}{
    \begin{tabular}{l ccc}
    \toprule
        Model size & Tiny & Small & Base \\
    \midrule\midrule
        Epochs & \multicolumn{3}{c}{130} \\
    \midrule
        Batch size & 1024 & 1024 & 400 \\
        Optimizer & \multicolumn{3}{c}{AdamW} \\
        Base Learning rate & 2e-4 & 2e-4 & 5e-5 \\
        Learning rate decay & \multicolumn{3}{c}{cosine} \\
        Weight decay & \multicolumn{3}{c}{1e-8}\\
        Warmup epochs & \multicolumn{3}{c}{5} \\
    \midrule
        Label smoothing & \multicolumn{3}{c}{0.1}\\
        Dropout & \multicolumn{3}{c}{\xmark} \\
        Stoch. Depth & 0.0 & 0.1 & 0.3 \\
        Repeated Aug & \multicolumn{3}{c}{\cmark} \\
        Gradient Clip. & \multicolumn{3}{c}{\cmark} \\
    \midrule
        Rand Augment & \multicolumn{3}{c}{9 / 0.5}\\
        Mixup prob. & \multicolumn{3}{c}{0.0} \\
        Cutmix prob. & \multicolumn{3}{c}{1.0} \\
        Erasing prob. & \multicolumn{3}{c}{0.25} \\ 
    \bottomrule
    \end{tabular}
    }
    \caption{\textbf{Settings of \ours on ImageNet-1K classification, fine-tuning from pre-trained weights.} We mostly follow the settings open to public by the authors~\cite{deit}.}
    \label{tab:in1k_settings_finetune}
    \end{minipage}
\end{table*}
\begin{table*}[t!]
    \centering
    \begin{minipage}{0.45\textwidth}
        \centering
        \resizebox{\textwidth}{!}{
            \begin{tabular}{l c c c}
                \toprule
                    Datasets & iNatuarlist-18, -19 & Flowers, StanfordCars  \\
                \midrule\midrule
                    Epochs & 300 & 1000 \\
                \midrule
                    Batch size & 768 & 768 \\
                    Optimizer & AdamW & AdamW \\
                    Base learning rate & 5e-5 & 1e-4 \\
                    Learning rate decay & cosine & cosine \\
                    Weight decay & 1e-6 & 5e-2 \\
                    Warmup epochs & 5 & 5 \\
                \midrule
                    Label smoothing & 0.1 & 0.1  \\
                    Dropout & \xmark & \xmark \\
                    Stoch. Depth & 0.5 & 0.5 \\
                    Repeated Aug & \cmark & \cmark \\
                    Gradient Clip. & \xmark & \xmark \\
                \midrule
                    Rand Augment & 9/0.5 & 9/0.5 \\
                    Mixup prob. & 0.8 & 0.8 \\
                    Cutmix prob. & 1.0 & 1.0 \\
                    Erasing prob. & 0.1 & 0 \\ 
                \bottomrule
            \end{tabular}
        }
        \caption{\textbf{Training settings for transfer learning ImageNet-1K pre-trained \ours.} We mostly follow the settings open to public by the authors~\cite{deit}.}
        \label{tab:transfer_learning_appendix}
    \end{minipage}%
    \hspace{1cm}
    \begin{minipage}{0.45\textwidth}
        \centering
        \resizebox{\textwidth}{!}{
            \begin{tabular}{l c c c}
            \toprule
                 Datasets & iNatuarlist-18, -19 & Flowers, StanfordCars  \\
            \midrule\midrule
                Epochs & 300 & 1000 \\
            \midrule
                Batch size & 768 & 768 \\
                Optimizer & AdamW & AdamW \\
                Base learning rate & 5e-5 & 1e-4 \\
                Learning rate decay & cosine & cosine \\
                Weight decay & 1e-4 & 1e-2 \\
                Warmup epochs & 5 & 5 \\
            \midrule
                Label smoothing & 0.1 & 0.1  \\
                Dropout & \xmark & \xmark \\
                Stoch. Depth & 0.1 & 0.5 \\
                Repeated Aug & \cmark & \cmark \\
                Gradient Clip. & \xmark & \xmark \\
            \midrule
                Rand Augment & 9/0.5 & 9/0.5 \\
                Mixup prob. & 0.8 & 0.8 \\
                Cutmix prob. & 1.0 & 1.0 \\
                Erasing prob. & 0.1 & 0.25 \\ 
            \bottomrule
            \end{tabular}
        }
        \caption{\textbf{Training settings for transfer learning DINO pre-trained \ours.} We mostly follow the settings open to public by the authors~\cite{dino}.}
        \label{tab:transfer_learning_dino_appendix}
    \end{minipage}
\end{table*}
\subsection{Self-Supervised Learning}
\label{appendix:selsup}
\begin{figure*}[h!]
    \centering
    \includegraphics[width=0.8\linewidth]{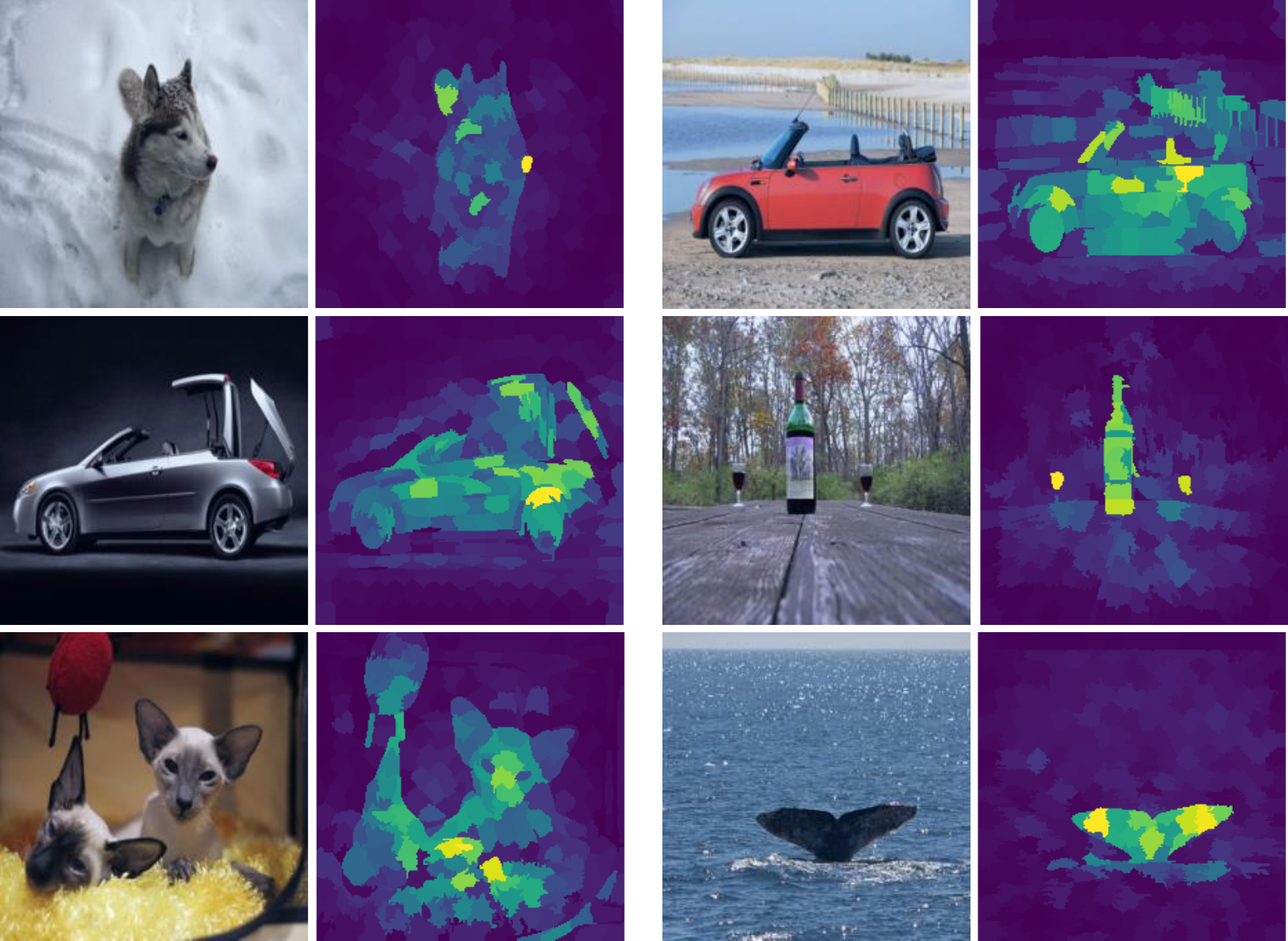} 
    \caption{\textbf{More qualitative results of the self-attention maps of DINO-\ours}. All maps are visualized by averaging the [CLS] token's self-attention from the final layer across all heads.}
    \label{fig:dino_attention_appendix} 
\end{figure*}

For DINO~\cite{dino} pre-training, we follow the exact training hyper-parameters used to pre-train ViTs, without any tuning, extra training techniques or augmentations. We train DINO-\ours for 300 epochs, and transfer the pre-trained model to downstream tasks following the settings mentioned in Section~\ref{sec:appendix_transfer}. Additional self-attention maps of DINO-\ours can be found in Figure~\ref{fig:dino_attention} and~\ref{fig:dino_attention_appendix}.

\begin{figure}[t!]
    \centering
    \includegraphics[width=\linewidth]{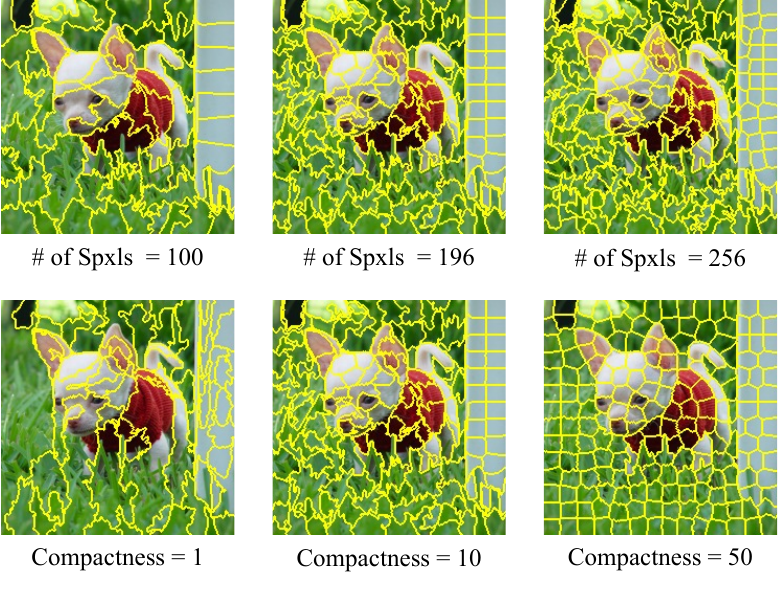}
    \caption{\textbf{Visulization of superpixel tokenization using various parameters.}}
    \label{fig:slic_vis_params}
\end{figure}

\subsection{Zero-shot Salient Object Segmentation}
\begin{figure*}[!t]
    \centering
    \includegraphics[width=0.7\linewidth]{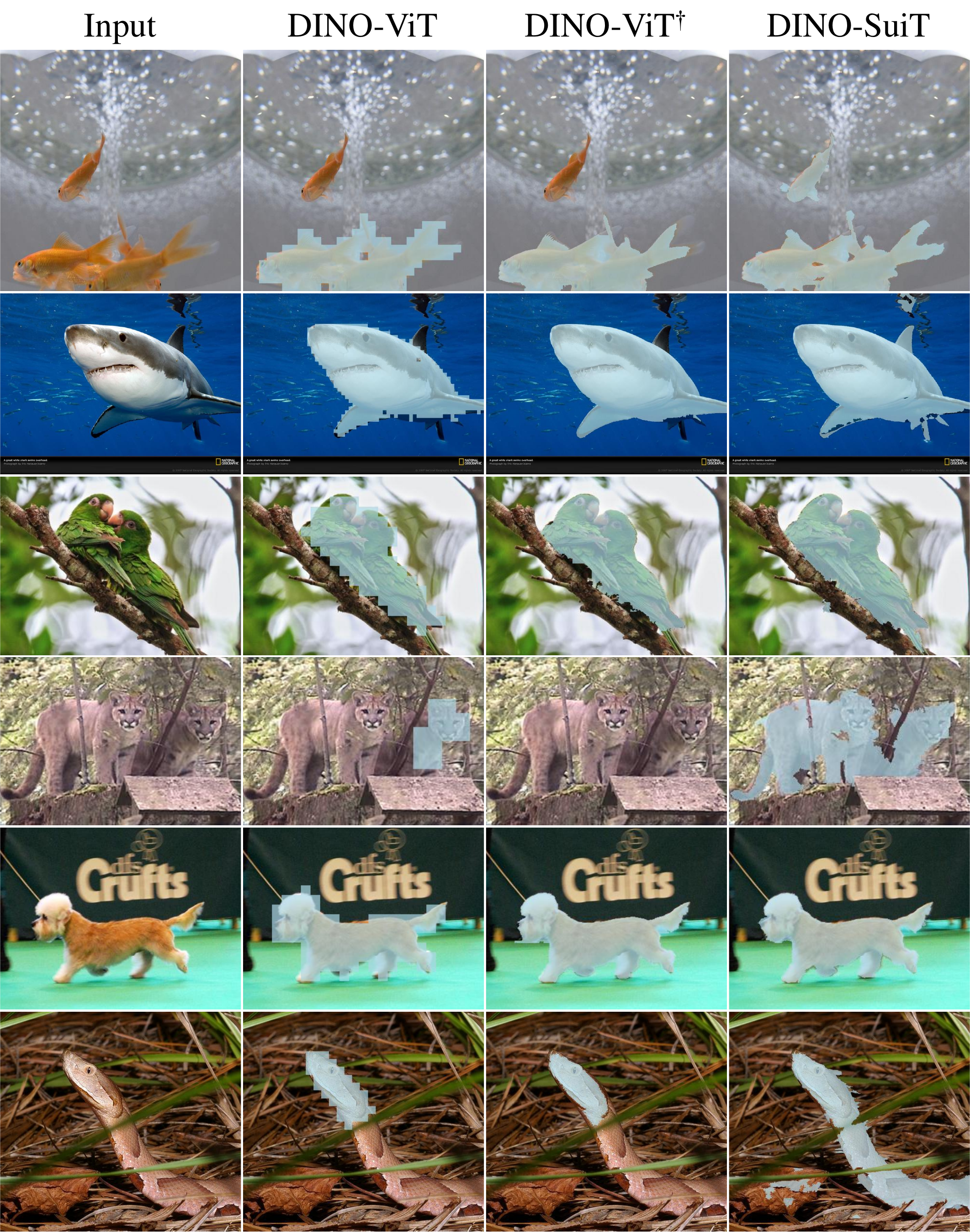} 
    \caption{\textbf{More qualitative results of zero-shot salient object segmentation}. $^{\dagger}$ denotes the model with additional post-processing.}
    \label{fig:zero_seg_appendix} 
\end{figure*}

In the TokenCut~\cite{tokencut} algorithm, we searched for and used the best $\tau$ value for  both DINO-ViT and DINO-\ours. For DINO-ViT, $\tau=0.2$ was used, while for DINO-\ours, $\tau=0.4$ was used. All other settings followed those described in the TokenCut~\cite{tokencut} paper. More qualitative results are shown in Figure~\ref{fig:zero_seg_appendix}.

\section{Semantic Integrity of Superpixel Tokens}
\label{sec:appendix_semantic_integrity}
\begin{figure*}[!t]
    \centering
    \includegraphics[width=0.8\linewidth]{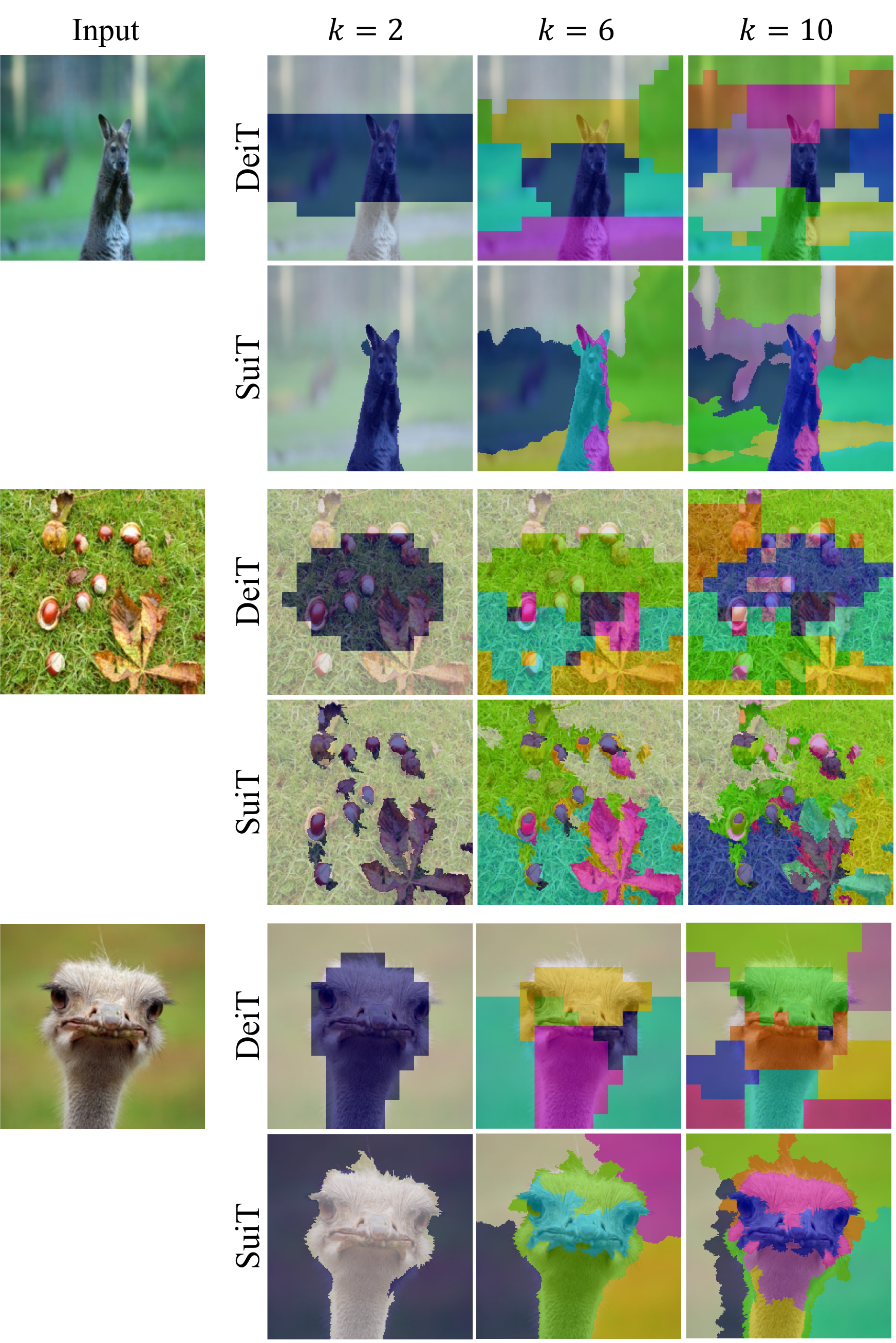} 
    \caption{\textbf{More qualitative results of semantic integrity of superpixel tokens with varying k}. For each input image, the first row represents the embeddings of DeiT tokens, while the second row represents the embeddings of \ours tokens.}
    \label{fig:kmeans_appendix} 
\end{figure*}

In our semantic integrity analysis, we performed $k$-means clustering on each token embeddings from DeiT~\cite{deit} and \ours. In Figure~\ref{fig:kmeans_appendix}, we visualized how tokens were clustered for each method by varying the value of $k$ from 2 to 10. This further shows that our superpixel tokens better preserve the semantics compared to that of DeiT. Most notably, when $k=2$, the tokens are clustered to two groups, foreground objects and background. As $k$ increases, the semantics within a cluster diverge and form clusters with more precise semantics.

\section{Visualization of Superpixel Tokenization}
We visualize how an image can be tokenized using various parameters in superpixel generation process at Figure~\ref{fig:slic_vis_params}. We show how the tokens change when setting different number of superpixels to be generated (compactness set to 10) and when setting different number of compactness (number of superpixels set to 196).

\end{document}